# HIGH DIMENSIONAL SEMIPARAMETRIC GAUSSIAN COPULA GRAPHICAL MODELS

By Han Liu[*], Fang Han, Ming Yuan,
John Lafferty, and Larry Wasserman

We propose a semiparametric approach called *the nonparanormal* SKEPTIC for efficiently and robustly estimating high dimensional undirected graphical models. To achieve modeling flexibility, we consider the nonparanormal graphical models proposed by Liu, Lafferty and Wasserman (2009). To achieve estimation robustness, we exploit nonparametric rank-based correlation coefficient estimators, including the Spearman's rho and Kendall's tau. We prove that the non-paranormal SKEPTIC achieves the optimal parametric rates of convergence for both graph recovery and parameter estimation. This result suggests that the nonparanormal graphical models can be used as a safe replacement of the popular Gaussian graphical models, even when the data are truly Gaussian. Besides theoretical analysis, we also conduct thorough numerical simulations to compare the graph recovery performance of different estimators under both ideal and noisy settings. The proposed methods are then applied on a large-scale genomic dataset to illustrate their empirical usefulness. The R package huge implementing the proposed methods is available on the Comprehensive R Archive Network: http://cran.r-project.org/.

**1. Introduction.** We consider the problem of estimating high dimensional undirected graphical models. Given $n$ independent observations from a $d$-dimensional random vector $X := (X_1, ..., X_d)^T$, we want to estimate an undirected graph $G := (V, E)$, where $V := \{1, \ldots, d\}$ contains nodes corresponding to the $d$ variables in $X$, and the edge set $E$ describes the conditional independence relationships between $X_1, \ldots, X_d$. Let $X_{\setminus\{i,j\}} := \{X_k : k \neq i, j\}$, we say the joint distribution of $X$ is Markov to $G$ if $X_i$ is independent of $X_j$ given $X_{\setminus\{i,j\}}$ for all $(i, j) \notin E$.

One popular method for this problem is the Gaussian graphical model, in which the random vector $X$ is assumed to be Gaussian: $X \sim N_d(\mu, \Sigma)$.

---

*The research of Han Liu, John Lafferty, and Larry Wasserman was supported by NSF grant IIS-1116730 and AFOSR contract FA9550-09-1-0373. We thank David Donoho for constructive comments and helpful discussions. We also thank Tuo Zhao for helpful discussions of this work.

Under this normality assumption, the graph $G$ is encoded by the precision matrix $\Omega := \Sigma^{-1}$. More specifically, no edge connects $X_j$ and $X_k$ if and only if $\Omega_{jk} = 0$ (Dempster, 1972). In low dimensions where $d < n$, Drton and Perlman (2007, 2008) develop a multiple testing procedure for identifying the sparsity pattern of the precision matrix. In high dimensions where $d \gg n$, Meinshausen and Bühlmann (2006) propose a neighborhood pursuit approach for estimating Gaussian graphical models by solving a collection of sparse regression problems using the Lasso in parallel. Yuan and Lin (2007), Banerjee, Ghaoui and d'Aspremont (2008), and Friedman, Hastie and Tibshirani (2008) develop a penalized likelihood approach to directly estimate $\Omega$. Rothman et al. (2008), Ravikumar et al. (2009a), and Lam and Fan (2009) study the theoretical properties of the penalized likelihood methods. More recently, Yuan (2010) and Cai, Liu and Luo (2011) propose the graphical Dantzig selector and CLIME, respectively. Both of these methods can be solved by linear programming and have more favorable theoretical properties than the penalized likelihood approach.

There are two drawbacks of the Gaussian graphical model: (i) the distributions of the data are in general non-Gaussian; (ii) the data could be noisy (e.g. rothamated by outliers). To handle the first challenge, Liu, Lafferty and Wasserman (2009) propose the *nonparanormal* family to relax the Gaussian assumption. A random vector $X$ belongs to a nonparanormal family if there exists a set of univariate monotone functions $\{f_j\}_{j=1}^d$ such that $f(X) := (f_1(X_1), \ldots, f_d(X_d))^T$ is Gaussian. They provide an estimation algorithm that has the same computational cost as the graphical lasso (glasso), while it achieves the rate of convergence $O(\sqrt{n^{-1/2} \log d})$ for estimating the precision matrix in the Frobenious and spectral norms. Other nonparametric graph estimation methods include forest graphical models or conditional graphical models Liu et al. (2011) and Liu et al. (2010).

In this paper we show that the rate of convergence obtained by Liu, Lafferty and Wasserman (2009) is not optimal. We present an alternative procedure that simultaneously achieves estimation robustness and rate optimality. The main idea is to exploit robust nonparametric rank-based statistics including Spearman's rho and Kendall's tau to directly estimate the unknown correlation matrix, without explicitly calculating the marginal transformations. We call this approach *nonparanormal* SKEPTIC (since the Spearman/Kendall estimates preempt transformations to infer correlation). The estimated correlation matrix is then plugged into existing parametric procedures (e.g., the graphical lasso, CLIME, or graphical Dantzig selector) to obtain the final estimate of the inverse correlation matrix and the graph.

By leveraging existing analysis (Cai, Liu and Luo, 2011; Lam and Fan,

2009; Ravikumar et al., 2009a; Yuan, 2010), we prove that although the nonparanormal family is larger than the Gaussian family, the nonparanormal SKEPTIC achieves the optimal parametric rates of convergence in terms of both precision matrix estimation and graph recovery. This result suggests that the extra modeling flexibility and robustness come at almost no cost in terms of statistical efficiency. Therefore, the nonparanormal SKEPTIC can be used as a safe replacement for Gaussian estimators even when the data are truly Gaussian. Moreover, by avoiding the estimation of the transformation functions, this new approach has fewer tuning parameters than the original method proposed by Liu, Lafferty and Wasserman (2009).

We provide thorough numerical studies to support our theory. Our results show that, when the data contamination rate is low, the normal-score based nonparanormal estimator proposed by Liu, Lafferty and Wasserman (2009) is slightly more efficient than the nonparanormal SKEPTIC. However, when the data contamination rate is higher, the nonparanormal SKEPTIC significantly outperforms the normal-score based estimator. This result reflects a tradeoff between statistical efficiency and estimation robustness.

In a related work, Xue and Zou (2012) independently proposed a similar regularized rank-based estimation idea for estimating nonparanormal graphical models. The main difference between our work and theirs is that Xue and Zou (2012) only propose the use of Spearman's rho estimator, while we study both Spearman's rho and Kendall's tau estimators. Another major difference is that the current paper compares the rank-based estimators with the normal-score based estimators and discusses their robustness properties, while Xue and Zou (2012) propose and analyze adaptive versions of rank-based Dantzig selector and CLIME estimators.

The rest of the paper is organized as follows. In the next section we briefly review some background on the nonparanormal estimator from Liu, Lafferty and Wasserman (2009). In Section 3 we present the nonparanormal SKEPTIC estimator, which exploits Spearman's rho and Kendall's tau statistics to estimate the underlying correlation matrix. Although not necessary for the SKEPTIC, we also provide results on consistently estimating the marginal transformations to Normality. In Section 4 we present a theoretical analysis of the method, with more detailed proofs collected in the appendix. In Section 5 we present numerical results on both simulated and real data, where the problem is to construct large undirected graphs for different biological entities (different tissue types or genes) using large-scale genomic datasets. We then discuss the connections to existing methods and possible future directions in the last section. Some of the results in this paper were first stated without proof in a conference version:

**2. Background.** We describe the nonparanormal family and the normal-score based estimator proposed by Liu, Lafferty and Wasserman (2009).

2.1. *Notation.* Let $A = [A_{jk}] \in \mathbb{R}^{d \times d}$ and $v = (v_1, \ldots, v_d)^T \in \mathbb{R}^d$. For $1 \leq q < \infty$, we define $\|v\|_q = \left(\sum_{i=1}^d |v_i|^q\right)^{1/q}$ and $\|v\|_\infty = \max_{1 \leq i \leq d} |v_i|$. The matrix $\ell_q$-operator norm is $\|A\|_q = \sup_{v \neq 0} \frac{\|Av\|_q}{\|v\|_q}$. In particular, for $q = 1$ and $q = \infty$, $\|A\|_1 = \max_{1 \leq j \leq d} \sum_{i=1}^d |A_{ij}|$ and $\|A\|_\infty = \max_{1 \leq i \leq d} \sum_{j=1}^d |A_{ij}|$. The matrix $\ell_2$-operator norm, or spectral norm, is the largest singular value. We denote $\|A\|_{\max} = \max_{j,k} |A_{jk}|$ and $\|A\|_F^2 = \sum_{j,k} |A_{jk}|^2$. We also denote $v_{\backslash j} = (v_1, \ldots, v_{j-1}, v_{j+1}, \ldots, v_d)^T \in \mathbb{R}^{d-1}$ and similarly denote by $A_{\backslash i, \backslash j}$ the submatrix of $A$ obtained by removing the $i^{\text{th}}$ row and $j^{\text{th}}$ column. We use $A_{i, \backslash j}$ to represent the $i^{\text{th}}$ row of $A$ with its $j^{\text{th}}$ entry removed. We use $\lambda_{\min}(A)$ and $\lambda_{\max}(A)$ to denote the smallest and largest eigenvalues of $A$.

2.2. *The Nonparanormal Distribution.* The nonparanormal family is a nonparametric extension of the Normal family. Using the same idea as sparse additive models (Liu, Lafferty and Wasserman, 2008; Liu and Zhang, 2009; Ravikumar et al., 2009b), we replace the random variable $X = (X_1, \ldots, X_d)^T$ by the transformed variable $f(X) = (f_1(X_1), \ldots, f_d(X_d))^T$, and assume that $f(X)$ is multivariate Gaussian. The nonparanormal only depends on the univariate functions $\{f_j\}_{j=1}^d$ and the correlation matrix $\Sigma^0$, all of which are to be estimated from data. More precisely, we have the following definition.

DEFINITION 2.1 (Nonparanormal). *Let $f = \{f_1, \ldots, f_d\}$ be a set of monotone univariate functions and let $\Sigma^0 \in \mathbb{R}^{d \times d}$ be a positive-definite correlation matrix with* $\text{diag}\left(\Sigma^0\right) = \mathbf{1}$*. We say a d-dimensional random variable $X = (X_1, \ldots, X_d)^T$ has a nonparanormal distribution $X \sim NPN_d(f, \Sigma^0)$ if $f(X) := (f_1(X_1), \ldots, f_d(X_d))^T \sim N_d(0, \Sigma^0)$.*

For continuous distributions, the nonparanormal family is equivalent to the Gaussian copula family (Klaassen and Wellner, 1997; Tsukahara, 2005).

Let $\Omega^0 = \left(\Sigma^0\right)^{-1}$ be the precision matrix. Liu, Lafferty and Wasserman (2009) prove that $\Omega^0$ encodes the undirected graph of $X$, i.e., $\Omega_{jk}^0 = 0 \Leftrightarrow X_j \perp\!\!\!\perp X_k \mid X_{\backslash\{j,k\}}$. Therefore, to estimate the graph for the nonparanormal family, it suffices to estimate the sparsity pattern of $\Omega^0$. More discussions can be found in Lafferty, Liu and Wasserman (2012).

2.3. *The Normal-score Estimator.* Liu, Lafferty and Wasserman (2009) suggest a two-step procedure to estimate the graph.

1. Replace the observations by their corresponding normal-scores.
2. Apply the glasso to the transformed data to estimate the graph.

More specifically, let $x^1, \ldots, x^n \in \mathbb{R}^d$ be $n$ data points and let $I(\cdot)$ be the indicator function. We define $\widehat{F}_j(t) = \frac{1}{n+1} \sum_{i=1}^n I(x_j^i \leq t)$ to be the scaled empirical cumulative distribution function of $X_j$. Liu, Lafferty and Wasserman (2009) study the estimator of the nonparanormal transformation functions given by[1] $\widehat{f}_j(t) = \Phi^{-1}\left(T_{\delta_n}[\widehat{F}_j(t)]\right)$, where $\Phi^{-1}(\cdot)$ is the standard Gaussian quantile function and $T_{\delta_n}$ is a Winsorization (or truncation) operator defined as $T_{\delta_n}(x) := \delta_n \cdot I(x < \delta_n) + x \cdot I(\delta_n \leq x \leq 1 - \delta_n) + (1 - \delta_n) \cdot I(x > 1 - \delta_n)$. Let $\widehat{S}^{\mathrm{ns}} = [\widehat{S}^{\mathrm{ns}}_{jk}]$ be the correlation matrix of the transformed data, where

$$(2.1) \qquad \widehat{S}^{\mathrm{ns}}_{jk} = \frac{\frac{1}{n} \sum_{i=1}^n \widehat{f}_j(x_j^i) \widehat{f}_k(x_k^i)}{\sqrt{\frac{1}{n} \sum_{i=1}^n \widehat{f}_j^2(x_j^i)} \cdot \sqrt{\frac{1}{n} \sum_{i=1}^n \widehat{f}_k^2(x_k^i)}}.$$

The nonparanormal estimate of the inverse correlation matrix $\widehat{\Omega}^{\mathrm{ns}}$ can be obtained by plugging $\widehat{S}^{\mathrm{ns}}$ into the glasso.

Taking $\delta_n = \frac{1}{n+1}$, we call $\widehat{S}^{\mathrm{ns}}_{jk}$ the *normal-score rank correlation coefficient.* For bivariate Gaussian copula distributions, Klaassen and Wellner (1997) prove that $\widehat{S}^{\mathrm{ns}}_{jk}$ is efficient in estimating $\Sigma^0_{jk}$. However, it appears that their efficiency result cannot be generalized to the high dimensional setting. The reason is that the standard Gaussian quantile function $\Phi^{-1}(\cdot)$ diverges very quickly when it is evaluated at a point close to 1. To handle high dimensional cases, Liu, Lafferty and Wasserman (2009) suggest to use a truncation level $\delta_n$. Such a truncation level $\delta_n = \frac{1}{4n^{1/4}\sqrt{\pi \log n}}$ is chosen to control the tradeoff of bias and variance in high dimensions. They analyzed the high dimensional scaling of the precision matrix estimator $\widehat{\Omega}^{\mathrm{ns}}$ and showed that

$$(2.2) \qquad \|\widehat{\Omega}^{\mathrm{ns}} - \Omega^0\|_F = O_P\left(\sqrt{\frac{(s+d)\log d + \log^2 n}{n^{1/2}}}\right),$$

$$(2.3) \qquad \|\widehat{\Omega}^{\mathrm{ns}} - \Omega^0\|_2 = O_P\left(\sqrt{\frac{s \log d + \log^2 n}{n^{1/2}}}\right),$$

where $s := \mathrm{Card}\left(\left\{(j,k) \in \{1, \ldots, d\} \times \{1, \ldots, d\} \,|\, \Omega^0_{jk} \neq 0, \ j \neq k\right\}\right)$ is the number of nonzero off-diagonal elements of the true precision matrix.

Using the results of Ravikumar et al. (2009a), it can also be shown that, under appropriate conditions, the sparsity pattern of the precision matrix can be accurately recovered with high probability. In particular, the nonparanormal estimator $\widehat{\Omega}^{\mathrm{ns}}$ satisfies $\mathbb{P}\left(\mathcal{G}\left(\widehat{\Omega}^{\mathrm{ns}}, \Omega^0\right)\right) \geq 1 - o(1)$, where $\mathcal{G}(\widehat{\Omega}^{\mathrm{ns}}, \Omega^0)$ is the event $\{\mathrm{sign}\big(\widehat{\Omega}_{jk}^{\mathrm{ns}}\big) = \mathrm{sign}(\Omega_{jk}^0), \forall j, k \in \{1, \ldots, d\}\}$. We refer to [Liu, Lafferty and Wasserman](#) (2009) for more details.

---

[1] Instead of $\widehat{F}_j$, Liu, Lafferty and Wasserman (2009) use the standard empirical cumulative distribution function. These two estimators are asymptotically equivalent.

In the next section, we show that the rates in (2.2) and (2.3) are not optimal and provide an alternative estimator that achieves the optimal rate.

## 3. The Nonparanormal SKEPTIC.

Nonparanormal distributions have two types of parameters: the precision matrix $\Omega^0 := (\Sigma^0)^{-1}$ and the marginal transformations $\{f_j\}_{j=1}^d$. In this section we develop methods for estimating both types of parameters. The main idea behind our new procedure is to exploit Spearman's rho and Kendall's tau statistics to directly estimate $\Omega^0$, without explicitly calculating the marginal transformation functions $\{f_j\}_{j=1}^d$. We then estimate the marginal transformations separately.

More specifically, let $r_j^i$ be the rank of $x_j^i$ among $x_j^1, \ldots, x_j^n$ and $\bar{r}_j = \frac{1}{n} \sum_{i=1}^n r_j^i = \frac{n+1}{2}$. We consider the following statistics:

(Spearman's rho)  $\widehat{\rho}_{jk} = \dfrac{\sum_{i=1}^n (r_j^i - \bar{r}_j)(r_k^i - \bar{r}_k)}{\sqrt{\sum_{i=1}^n (r_j^i - \bar{r}_j)^2 \cdot \sum_{i=1}^n (r_k^i - \bar{r}_k)^2}},$

(Kendall's tau)  $\widehat{\tau}_{jk} = \dfrac{2}{n(n-1)} \sum_{1 \leq i < i' \leq n} \mathrm{sign}\left(\big(x_j^i - x_j^{i'}\big)\big(x_k^i - x_k^{i'}\big)\right).$

Both $\widehat{\rho}_{jk}$ and $\widehat{\tau}_{jk}$ are nonparametric correlations between the empirical realizations of random variables $X_j$ and $X_k$. Note that these statistics are invariant under monotone transformations. For Gaussian random variables there is a one-to-one mapping between these two statistics; details can be found in [Kendall](#) (1948) and [Kruskal](#) (1958). Let $\widetilde{X}_j$ and $\widetilde{X}_k$ be two independent copies of $X_j$ and $X_k$. We denote by $F_j$ and $F_k$ the CDFs of $X_j$ and $X_k$. The population versions of Spearman's rho and Kendall's tau are given by $\rho_{jk} := \mathrm{Corr}\big(F_j(X_j), F_k(X_k)\big)$ and $\tau_{jk} := \mathrm{Corr}\big(\mathrm{sign}(X_j - \widetilde{X}_j), \mathrm{sign}(X_k - \widetilde{X}_k)\big)$. Both $\rho_{jk}$ and $\tau_{jk}$ are association measures based on the notion of concordance. We call two pairs of real numbers $(s, t)$ and $(u, v)$ concordant if $(s - t)(u - v) > 0$ and disconcordant if $(s - t)(u - v) < 0$. The following proposition provides further insight into the relationship between $\rho_{jk}$ and $\tau_{jk}$. The proof is provided in the appendix for completeness.

PROPOSITION 3.1. *Let $(X_j^{(1)}, X_k^{(1)}), (X_j^{(2)}, X_k^{(2)})$, and $(X_j^{(3)}, X_k^{(3)})$ be three independent random vectors with the same distribution as $(X_j, X_k)$. Define*

$$\mathsf{C}(j, s, t; k, u, v) = \mathbb{P}\big((X_j^{(s)} - X_j^{(t)})(X_k^{(u)} - X_k^{(v)}) > 0\big),$$

$$\mathsf{D}(j,s,t;k,u,v) = \mathbb{P}\big((X_j^{(s)} - X_j^{(t)})(X_k^{(u)} - X_k^{(v)}) < 0\big).$$

*Then $\rho_{jk} = 3\mathsf{C}(j,1,2;k,1,3) - 3\mathsf{D}(j,1,2;k,1,3)$ and $\tau_{jk} = \mathsf{C}(j,1,2;k,1,2) - \mathsf{D}(j,1,2;k,1,2)$.*

For nonparanormal distributions, the following lemma connects Spearman's rho and Kendall's tau to the underlying Pearson correlation coefficient $\Sigma_{jk}^0$.

LEMMA 3.1 (Kendall (1948); Kruskal (1958)). *Assuming $X \sim NPN_d(f, \Sigma^0)$, we have $\Sigma_{jk}^0 = 2\sin\left(\frac{\pi}{6}\rho_{jk}\right) = \sin\left(\frac{\pi}{2}\tau_{jk}\right)$.*

Motivated by this lemma, we define the following estimators $\widehat{S}^\rho = [\widehat{S}_{jk}^\rho]$ and $\widehat{S}^\tau = [\widehat{S}_{jk}^\tau]$ for the unknown correlation matrix $\Sigma^0$:

$$\widehat{S}_{jk}^\rho = \left\{ \begin{array}{ll} 2\sin\left(\dfrac{\pi}{6}\widehat{\rho}_{jk}\right) & j \neq k \\ 1 & j = k \end{array} \right. \quad \text{and} \quad \widehat{S}_{jk}^\tau = \left\{ \begin{array}{ll} \sin\left(\dfrac{\pi}{2}\widehat{\tau}_{jk}\right) & j \neq k \\ 1 & j = k \end{array} \right. .$$

As will be shown in later sections, the final graph estimators based on Spearman's rho and Kendall's tau statistics have similar theoretical performance. Thus, in the following sections we omit the superscripts $\rho$ and $\tau$ and simply denote the estimated correlation matrix by $\widehat{S}$.

3.1. *Estimating Sparse Precision Matrices and Graphs.* In this subsection, we explain how to exploit the estimated correlation matrices $\widehat{S}^\tau$ and $\widehat{S}^\rho$ to estimate the sparse precision matrix and graph.

3.1.1. *The Nonparanormal* SKEPTIC *with the Graphical Dantzig Selector.* The main idea of the graphical Dantzig selector (Yuan, 2010) is to take advantage of the connection between multivariate linear regression and entries of the inverse covariance matrix. The following is the detailed algorithm, where $\delta$ is a tuning parameter.

- Estimation: For $j = 1, \ldots, d$, calculate

(3.1) $$\widehat{\theta}^j = \operatorname*{arg\,min}_{\theta \in \mathbb{R}^{d-1}} \|\theta\|_1 \text{ subject to } \|\widehat{S}_{\setminus j,j} - \widehat{S}_{\setminus j,\setminus j}\theta\|_\infty \leq \delta,$$

$$\widehat{\Omega}_{jj} = \left[1 - 2(\widehat{\theta}^j)^T \widehat{S}_{\setminus j,j} + (\widehat{\theta}^j)^T \widehat{S}_{\setminus j,\setminus j}\widehat{\theta}^j\right] \text{ and } \widehat{\Omega}_{\setminus j,j} = -\widehat{\Omega}_{jj}\widehat{\theta}^j.$$

- Symmetrization:

(3.2) $$\widehat{\Omega}^{\mathrm{gDS}} = \operatorname*{arg\,min}_{\Omega = \Omega^T} \|\Omega - \widehat{\Omega}\|_1.$$

Within each iteration, the Dantzig selector in (3.1) can be formulated as a linear program. A more sophisticated path algorithm (DASSO) to solve the Dantzig selector has been developed by James, Radchenko and Lv (2009).

3.1.2. *The Nonparanormal* SKEPTIC *with CLIME.* Let $\mathbf{I}_d$ be the $d$-dimensional identity matrix. The estimated correlation coefficient matrix $\widehat{S}$ can also be plugged into the CLIME estimator (Cai, Liu and Luo, 2011), which is defined by

$$(3.3) \quad \widehat{\Omega}^{\text{CLIME}} = \arg\min_{\Omega} \sum_{j,k} |\Omega_{jk}| \text{ subject to } \|\widehat{S}\Omega - \mathbf{I}_d\|_{\max} \leq \Delta,$$

where $\Delta$ is the tuning parameter. Cai, Liu and Luo (2011) show that this convex optimization can be decomposed into $d$ vector minimization problems, each of which can be cast as a linear program. Thus CLIME has the potential to scale to large datasets.

3.1.3. *The Nonparanormal* SKEPTIC *with the Graphical Lasso.* We can also plug the estimated correlation matrix $\widehat{S}$ into the graphical lasso:

$$(3.4) \qquad \widehat{\Omega}^{\text{glasso}} = \arg\min_{\Omega \succ 0} \left\{ \text{tr}\left(\widehat{S}\Omega\right) - \log|\Omega| + \lambda \sum_{j,k} |\Omega_{jk}| \right\}.$$

One thing to note is that $\widehat{S}$ may not be positive semidefinite. Even though the formulation (3.4) is still convex, certain algorithms (like the blockwise-coordinate descent algorithm (Friedman, Hastie and Tibshirani, 2008)) may fail if $\widehat{S}$ is indefinite. However other algorithms like the two-metric projected Newton method or first-order projection do not have such positive semidefinite assumption on $\widehat{S}$. These algorithms can be directly exploited to efficiently solve (3.4).

Unlike the graphical Lasso formulation, the graphical Dantzig selector and CLIME can both be formulated as linear programs, so they do not require positive semidefiniteness of the input correlation matrix.

3.1.4. *The Nonparanormal* SKEPTIC *with the Neighborhood pursuit Estimator (The Meinshausen-Bühlmann procedure).* The nonparanormal SKEPTIC can also be applied with the Meinshausen-Bühlmann procedure to estimate the graph. As has been discussed in Friedman, Hastie and Tibshirani (2008), the correlation matrix is a sufficient statistic for the Meinshausen-Bühlmann procedure. However, in this case, we need to make sure that $\widehat{S}$ is positive semidefinite. Otherwise, the algorithm may not converge. Practically, we can first project $\widehat{S}$ into the cone of positive semidefinite matrices. In particular, we need to solve the following convex optimization problem:

$$(3.5) \qquad \widetilde{S} = \arg\min_{S \succeq 0} \|\widehat{S} - S\|_{\max}.$$

Here we use the $\|\cdot\|_{\max}$-norm instead of the $\|\cdot\|_F$-norm , due to theoretical concerns developed in the next section. In fact, the optimization problem in (3.5) can be formulated as the dual of a graphical lasso problem. To find the projection solution, we need to search for the smallest possible tuning parameter which still makes the optimization problem feasible. Empirically, we can use a surrogate projection procedure that computes a singular value decomposition of $\widehat{S}$ and truncates all of the negative singular values to be zero.

3.2. *Computational Complexity.* Compared to the corresponding parametric methods like the graphical lasso, graphical Dantzig selector, CLIME, and the Meinshausen-Bühlmann estimator, the only extra cost of the nonparanormal SKEPTIC is the computation of $\widehat{S}$, which requires us to calculate the $d(d-1)/2$ pairs of Spearman's rho or Kendall's tau statistics. A naive implementation of Kendall's tau statistic requires $O(n^2)$ flops. However, efficient algorithm based on sorting and balanced binary trees has been developed to calculate Kendall's tau statistic with complexity $O(n \log n)$. Details can be found in Christensen (2005). The computation of Spearman's rho statistic only requires one sort of the data, which has complexity $O(n \log n)$.

3.3. *Estimating Marginal Transformations.* Though estimating the graph does not require estimating the marginal transformations, we are still interested in estimating marginal transformations. Estimating marginal transformations is useful for calculating the likelihood of a nonparanormal fit. Let $\widetilde{F}_j(t) = \frac{1}{n} \sum_{i=1}^n I(x_j^i \leq t)$ be the empirical distribution function of $X_j$. We estimate the marginal transformation $f_j$ using the following estimator:

$$(3.6) \qquad \widetilde{f}_j(x) := \Phi^{-1}\left(T_{1/(2n)}\big[\widetilde{F}_j(x)\big]\right),$$

where the function $T_{\delta_n}(x) := \delta_n \cdot I(x < \delta_n) + x \cdot I(\delta_n \leq x \leq 1 - \delta_n) + (1 - \delta_n) \cdot I(x > 1 - \delta_n)$. An analysis of this estimator is given in the next section.

4. **Theoretical Properties.** We analyze the statistical properties of the nonparanormal SKEPTIC estimator. Our main result shows that $\widehat{S}^\rho$ and $\widehat{S}^\tau$ have a fast exponential concentration rate to $\Sigma^0$ in the $\|\cdot\|_{\max}$ norm. This result allows us to leverage existing analysis of different parametric methods to analyze the nonparanormal SKEPTIC estimator.

In particular, Theorem 4.3 states that the nonparanormal SKEPTIC achieves the same graph recovery and parameter estimation performance as the corresponding parametric methods. Since the nonparanormal family is much richer than the Gaussian family, such a result suggests that the nonparanormal SKEPTIC could be a safe replacement for Gaussian graphical models. We then use the graphical Dantzig selector as an illustrate example to showcase this result. Similar analysis can be carried on for both CLIME and the graphical lasso.

4.1. *Concentration Properties of the Estimated Correlation Matrices.* We first prove the concentration properties of the estimators $\widehat{S}^\rho$ and $\widehat{S}^\tau$. Let $\Sigma^0_{jk}$ be the Pearson correlation coefficient between $f_j(X_j)$ and $f_k(X_k)$. In terms of $\|\cdot\|_{\max}$ norm, we show that both $\widehat{S}^\rho$ and $\widehat{S}^\tau$ converge to $\Sigma^0$ in probability with the optimal parametric rate. Our results are based on different versions of Hoeffding's inequalities for U-statistics. Without loss of generality, in this paper we always assume $d > n$. The results for $d < n$ are straightforward.

THEOREM 4.1. *For any $n \geq \frac{21}{\log d} + 2$, with probability at least $1 - 1/d^2$, we have*

$$(4.1) \qquad \sup_{jk} \left| \widehat{S}^\rho_{jk} - \Sigma^0_{jk} \right| \leq 8\pi \sqrt{\frac{\log d}{n}}.$$

The next theorem illustrates the concentration property of $\widehat{S}^\tau$.

THEOREM 4.2. *For any $n > 1$, with probability at least $1 - 1/d$, we have*

$$(4.2) \qquad \sup_{jk} \left| \widehat{S}^\tau_{jk} - \Sigma^0_{jk} \right| \leq 2.45\pi \sqrt{\frac{\log d}{n}}.$$

With the above results we present the following "metatheorem," which shows that even though the nonparanormal SKEPTIC is a semiparametric estimator, it achieves the optimal parametric rate in high dimensions.

THEOREM 4.3 (Main Theorem). *If we plug the estimated matrix $\widehat{S}^\rho$ or $\widehat{S}^\tau$ into the parametric graphical lasso (or the graphical Dantzig selector, or CLIME), then under the same conditions on $\Sigma^0$ that ensure the consistency and graph recovery of these parametric methods under the Gaussian model, the nonparanormal SKEPTIC achieves the same (parametric) rate of convergence for both precision matrix estimation and graph recovery under the nonparanormal model.*

PROOF. The proof is based on the observation that the sample correlation matrix $\widehat{S}$ is a sufficient statistic for all three methods: the graphical lasso, graphical Dantzig selector, and CLIME. By examining the analysis in Cai, Liu and Luo (2011); Ravikumar et al. (2009a); Yuan (2010), a sufficient condition on $\widehat{S}$ to enable their analysis is that, there exists some constant $c$, such that $\mathbb{P}\left(\|\widehat{S} - \Sigma^0\|_{\max} > c\sqrt{\frac{\log d}{n}}\right) \leq 1 - \frac{1}{d}$, which can be replaced by (4.1) and (4.2) from theorems 4.1 and 4.2. □

The graphical lasso, graphical Dantzig selector, and CLIME have been proved to be minimax rate optimal over certain parameter classes under the Gaussian model. Since the nonparanormal family is strictly larger than the Gaussian family, we immediately justify the minimax optimality of the nonparanormal SKEPTIC estimator:

COROLLARY 4.1. *Over all the parameter spaces of $\Sigma^0$ such that the graphical lasso, graphical Dantzig, or CLIME are rate optimal under Gaussian models, the corresponding nonparanormal SKEPTIC estimator is also rate optimal for the same space of $\Sigma^0$ under the nonparanormal model.*

In terms of rates of convergence, the nonparanormal SKEPTIC can be a safe replacement of the Gaussian graphical models. The extra flexibility and robustness come at almost no cost. In the next subsection, we showcase this main theorem using the graphical Dantzig selector.

REMARK 4.1. *Even though in this section we only present the results on the graphical Dantzig selector, graphical lasso, and CLIME, similar arguments should hold for almost all methods that use the correlation matrix $\Sigma^0$ as a sufficient statistic.*

4.2. *Applying the Nonparanormal SKEPTIC with the graphical Dantzig Selector.* In Theorem 4.3 we have shown that the nonparanormal SKEPTIC estimator $\widehat{S}$ can be plugged into any parametric procedure and achieves the optimal parametric rate of convergence. In this subsection we use the graphical Dantzig selector as an example to see how this theorem can be applied in specific applications.

We denote $\widehat{\Omega}^{\text{npn-s}}$ to be the inverse correlation matrix estimated using the nonparanormal SKEPTIC with the graphical Dantzig selector in (3.2). Given a matrix $\Omega$, we define $\deg(\Omega) = \max_{1 \leq i \leq d} \sum_{j=1}^{d} I(|\Omega_{ij}| \neq 0)$. Following Yuan (2010), we consider a class of inverse correlation matrices: $\mathcal{M}_1(\kappa, \tau, M) := \Big\{ \Omega : \Omega \succ \mathbf{0}, \operatorname{diag}(\Omega^{-1}) = \mathbf{1}, \|\Omega\|_1 \leq \kappa, \frac{1}{\tau} \leq \lambda_{\min}(\Omega) \leq$

$\lambda_{\max}(\Omega) \leq \tau, \deg(\Omega) \leq M\Big\}$, where $\kappa, \tau > 1$. We then have the following corollary of Theorem 4.3.

THEOREM 4.4. *For $1 \leq q \leq \infty$, there exists a constant $C_1$ that depends on $\kappa$, $\tau$, $\lambda_{\min}(\Omega^0)$, and $\lambda_{\max}(\Omega^0)$, such that*

$$\sup_{\Omega^0 \in \mathcal{M}_1(\kappa, \tau, M)} \|\widehat{\Omega}^{\text{npn-s}} - \Omega^0\|_q = O_P\left(M\sqrt{\frac{\log d}{n}}\right),$$

*provided that $\lim_{n \to \infty} \dfrac{n}{M^2 \log d} = \infty$ and $\delta = C_1\sqrt{\dfrac{\log d}{n}}$, for sufficiently large $C_1$. Here $\delta$ is the tuning parameter used in (3.1).*

PROOF. The proof can be directly obtained by replacing Lemma 12 in Yuan (2010) with the result of Theorem 4.3.                                    □

The next theorem establishes the minimax lower bound for inverse correlation matrix estimation over the class $\mathcal{M}_1(\kappa, \tau, M)$. Its proof can be easily obtained by a modification of Theorem 5 in Yuan (2010).

THEOREM 4.5 (Yuan (2010)). *Let $M(\log d/n)^{1/2} = o(1)$. Then there exists a constant $C > 0$ depending only on $\kappa$ and $\tau$ such that*

$$\liminf_{n \to \infty} \inf_{\widehat{\Omega}} \sup_{\Omega^0 \in \mathcal{M}_1(\kappa, \tau, M)} \mathbb{P}\left(\|\widehat{\Omega} - \Omega^0\|_1 \geq CM\sqrt{\frac{\log d}{n}}\right) > 0,$$

*where the infimum is taken over all estimates of $\Omega$ based on the observed data $x^1, \ldots, x^n$.*

From the above theorems, we see that the nonparanormal SKEPTIC estimator of the inverse correlation matrix can achieve the parametric rate and is in fact minimax rate optimal over the parameter space $\mathcal{M}_1(\kappa, \tau, M)$ in terms of $\ell_1$-risk.

4.3. *Estimating Marginal Transformations.* Recall the definition of $\widetilde{f}_j(t)$ in (3.6). For any fixed $t$, $\widetilde{f}_j(t)$ converges in probability to $f_j(t)$. Theorem 4.6 provides a stronger result that $\widetilde{f}_j$ converges to $f_j$ uniformly over an expanding interval. This result is important for many downstream applications of nonparanormal modeling, e.g., discriminant analysis or principle component analysis; details will be provided in a follow-up paper.

THEOREM 4.6. *Let $g_j := f_j^{-1}$ be the inverse function of $f_j$. For any $0 < \gamma < 1$, define $I_n := \left[ g_j \left( -\sqrt{\frac{7}{4}(1-\gamma)\log n} \right), g_j \left( \sqrt{\frac{7}{4}(1-\gamma)\log n} \right) \right]$. Then $\sup\limits_{t \in I_n} \left| \widetilde{f}_j(t) - f_j(t) \right| = o_P(1)$.*

## 5. Experimental Results.

We investigate the empirical performance of different graph estimation methods on both synthetic and real datasets. In particular we consider the following methods: (i) npn – the original nonparanormal estimator from Liu, Lafferty and Wasserman (2009); (ii) normal – the Gaussian graphical model (which relies on the Gaussian assumption); (iii) npn-spearman – the nonparanormal SKEPTIC using Spearman's rho; (iv) npn-tau – the nonparanormal SKEPTIC using Kendall's tau; (v) npn-ns – the normal-score based estimator defined in (2.1) with $\delta_n = \frac{1}{n+1}$.

5.1. *Summary of the Results.* To compare the graph estimation performance of two procedures $A$ and $B$, in the following we use $A >_{\text{slightly}} B$ to represent that $A$ slightly outperforms $B$; $A > B$ means that $A$ is better than $B$; $A \gg B$ means that $A$ is significantly better than $B$; while $A \approx B$ means that $A$ and $B$ have similar performance. Here we summarize the main results:

- non-Gaussian data without outliers: npn-ns $\approx$ npn $\approx$ npn-spearman $\approx$ npn-tau $\gg$ normal.
- non-Gaussian data with a low level of outliers: npn-tau $\approx$ npn-spearman $>$ npn $>$ npn-ns $\gg$ normal.
- non-Gaussian data with a higher level of outliers: npn-tau $>$ npn-spearman $\gg$ npn $>$ npn-ns $\gg$ normal.
- Gaussian data without outliers: normal $\approx$ npn-ns $\approx$ npn $>_{\text{slightly}}$ npn-spearman $\approx$ npn-tau.
- Gaussian data with a low level of outliers: npn-tau $\approx$ npn-spearman $>$ npn $>$ npn-ns $\gg$ normal.
- Gaussian data with a higher level of outliers: npn-tau $>$ npn-spearman $>$ npn $>$ npn-ns $>$ normal.

These results indicate a tradeoff between estimation robustness and statistical efficiency. For nonparanormal data without outliers, npn-ns and npn behave similarly to npn-tau and npn-spearman. However, if the data are contaminated by outliers, npn-tau and npn-spearman outperform npn-ns and npn even when the contamination level is low. Overall, our simulations suggest that both npn-tau and npn-spearman have a good balance of statistical efficiency and robustness. In addition, since both the nonparanormal SKEPTIC and the normal-score based methods are rank-based, they are invariant to different choices of marginal transformations $f_j$ in the true model. In contrast, the Gaussian estimators (the graphical Lasso, CLIME, etc.) are not marginal transformation invariant. Their performance decreases dramatically when non-identity transformations are applied. Going beyond numerical simulations, we also apply our method to a large-scale genomic dataset.

5.2. *Numerical Simulations.* We adopt the same data generating procedure as in Liu, Lafferty and Wasserman (2009). To generate a $d$-dimensional sparse graph $G = (V, E)$, let $V = \{1, \ldots, d\}$ correspond to variables $X = (X_1, \ldots, X_d)$. We associate each index $j \in \{1, \ldots, d\}$ with a bivariate data point $(Y_j^{(1)}, Y_j^{(2)}) \in [0, 1]^2$ where $Y_1^{(k)}, \ldots, Y_d^{(k)} \sim \text{Uniform}[0, 1]$ for $k = 1, 2$. Each pair of vertices $(i, j)$ is included in the edge set $E$ with probability $\mathbb{P}\big((i, j) \in E\big) = \frac{1}{\sqrt{2\pi}} \exp\left(-\frac{\|y_i - y_j\|^2}{2s}\right)$ where $y_i := (y_i^{(1)}, y_i^{(2)})$ is the empirical observation of $(Y_i^{(1)}, Y_i^{(2)})$ and $\|\cdot\|$ denotes Euclidean distance. Here, $s = 0.125$ is a parameter that controls the sparsity level of the generated graph. We restrict the maximum degree of the graph to be 4 and build the inverse correlation matrix $\Omega^0$ according to $\Omega_{jk}^0 = 1$ if $j = k$, $\Omega_{jk}^0 = 0.245$ if $(j, k) \in E$, and $\Omega_{jk}^0 = 0$ otherwise. Here the value 0.245 guarantees positive definiteness of $\Omega^0$. Let $\Sigma^0 = (\Omega^0)^{-1}$. To obtain the correlation matrix, we simply rescale $\Sigma^0$ so that all its diagonal elements are 1. We then sample $n$ data points $x^1, \ldots, x^n$ from the nonparanormal distribution $NPN_d(f^0, \Sigma^0)$, where for simplicity we use the same univariate transformations on each dimension, i.e., $f_1^0 = \ldots = f_d^0 = f^0$. To sample data from the nonparanormal distribution, we also need $g^0 := (f^0)^{-1}$. The following two different versions of $g^0$ are used in the simulations:

DEFINITION 5.1. (Gaussian CDF Transformation) *Let $g_0$ be a univariate Gaussian cumulative distribution function with mean $\mu_{g_0}$ and the standard deviation $\sigma_{g_0}$: $g_0(t) := \Phi\left(\frac{t - \mu_{g_0}}{\sigma_{g_0}}\right)$. The Gaussian CDF transformation $g_j^0 = (f_j^0)^{-1}$ for the $j$-th dimension is defined as*

$$(5.1) \quad g_j^0(z_j) := \frac{g_0(z_j) - \int g_0(t)\phi\left(\frac{t - \mu_j}{\sigma_j}\right)dt}{\sqrt{\int \left(g_0(y) - \int g_0(t)\phi\left(\frac{t - \mu_j}{\sigma_j}\right)dt\right)^2 \phi\left(\frac{y - \mu_j}{\sigma_j}\right)dy}},$$

*where $\phi(\cdot)$ is the standard Gaussian density function.*

DEFINITION 5.2. (*Power Transformation*) *Let* $g_0(t) := \text{sign}(t)|t|^\alpha$ *where* $\alpha > 0$ *is a parameter. The power transformation for the $j$-th dimension is defined as*

$$(5.2) \qquad g_j^0(z_j) := \frac{g_0(z_j - \mu_j)}{\sqrt{\int g_0^2(t - \mu_j) \phi\left(\frac{t - \mu_j}{\sigma_j}\right) dt}},$$

*where* $\phi(\cdot)$ *is the standard Gaussian density function.*

These transformations were used by Liu, Lafferty and Wasserman (2009) to study the performance of the original nonparanormal estimator. To comply with their simulation design, for the Gaussian CDF transformation we set $\mu_{g_0} = 0.05$ and $\sigma_{g_0} = 0.4$; for the power transformation, we set $\alpha = 3$.

To generate synthetic data, we set $d = 100$, resulting in $\binom{100}{2} + 100 = 5,050$ parameters to be estimated. The sample sizes vary from $n = 100, 200$ to $500$. Three conditions are considered, corresponding to using the power transformation, the Gaussian CDF transformation, and linear transformation (or no transformation)[2].

To evaluate the robustness of these methods, we consider two types of data contamination mechanisms, *deterministic contamination* and *random contamination*. Let $r \in (0, 1)$ be the contamination level. For deterministic contamination we replace $\lfloor nr \rfloor$ data points with a deterministic vector $(+5, -5, +5, -5, +5, \ldots)^T \in \mathbb{R}^d$, in which the numbers $+5$ and $-5$ occur in an alternating way. For random contamination, we randomly (according to a uniform distribution) select $\lfloor nr \rfloor$ entries of each dimension and replace them with either $+5$ or $-5$ with equal probability. From the robustness point of view, the deterministic contamination is more malicious and can severely hurt non-robust procedures. In contrast, the random contamination is relatively benign and is more realistic for modern scientific data analysis.

Both the normal-score based nonparanormal estimators (npn and npn-ns) and the nonparanormal SKEPTIC estimators (npn-spearman and npn-tau) are two-step procedures. In the first step we obtain an estimate $\widehat{S}$ of the correlation matrix; in the second step we plug $\widehat{S}$ into a parametric graph estimation procedure. In this numerical study, we consider two parametric baseline procedures: (i) the graphical lasso and (ii) the Meinshausen-Bühlmann graph estimator. The former represents the likelihood-based approach and the latter is a type of pseudo-likelihood-based approach. In our experiments, we find that CLIME has behavior similar to the graphical lasso, while the graphical Dantzig selector behaves similar to the Meinshausen-Bühlmann method.

---

[2]For linear transformation, the data exactly follow the Gaussian distribution.

Our implementations of the nonparanormal SKEPTIC, graphical lasso and Meinshausen-Bühlmann methods are available in the R package huge[3].

Let $G = (V, E)$ be a $d$-dimensional graph. We denote by $|E|$ the number of edges in the graph $G$. We use false positive and negative rates to evaluate the graph estimation performance. Let $\widehat{G}^\lambda = (V, \widehat{E}^\lambda)$ be an estimated graph using the regularization parameter $\lambda$ in either the graphical lasso procedure (3.4) or the Meinshausen-Bühlmann procedure. The number of false positives when using the regularization parameter $\lambda$ is $\mathrm{FP}(\lambda) :=$ the number of edges in $\widehat{E}^\lambda$ but not in $E$. The number of false negatives at $\lambda$ is defined as $\mathrm{FN}(\lambda) :=$ the number of edges in $E$ but not in $\widehat{E}^\lambda$. We further define the false negative rate (FNR) and false positive rate (FPR) as

$$(5.3) \quad \mathrm{FNR}(\lambda) := \frac{\mathrm{FN}(\lambda)}{|E|} \quad \text{and} \quad \mathrm{FPR}(\lambda) := \mathrm{FP}(\lambda) / \left[ \binom{d}{2} - |E| \right].$$

Let $\Lambda$ be the set of all regularization parameters used to create the full path. The oracle regularization parameter $\lambda^*$ is defined as

$$\lambda^* := \arg\min_{\lambda \in \Lambda} \left\{ \mathrm{FNR}(\lambda) + \mathrm{FPR}(\lambda) \right\}.$$

The oracle score is defined to be $\mathrm{FNR}(\lambda^*) + \mathrm{FPR}(\lambda^*)$. To illustrate the overall performance of the studied methods over the full paths, the averaged ROC curves for $n = 200, d = 100$ over 100 trials are shown in Figures 1 to 4, using $(\mathrm{FNR}(\lambda), 1 - \mathrm{FPR}(\lambda))$. For each figure five curves are presented, corresponding to npn, npn-tau, npn-spearman, npn-ns, and normal.

Let FPR := $\mathrm{FPR}(\lambda^*)$ and FNR := $\mathrm{FNR}(\lambda^*)$. Figures 5 to 8 provide numerical comparisons of the three methods on datasets with the different transformations, where we repeat the experiments 100 times and report the average FPR and FNR values with the corresponding standard errors in parentheses.

In the following we provide detailed analysis based on these numerical simulations.

5.2.1. *Non-Gaussian Data with No Outliers.* From the power transformation and CDF transformation plots in Figures 1 to 4, we see that, when the contamination level $r$ is zero, the performance of the nonparanormal SKEPTIC estimators (npn-spearman and npn-tau) and the previous normal-score based nonparanormal estimators (npn, and npn-ns) are comparable. In this case, all these methods significantly outperform the corresponding

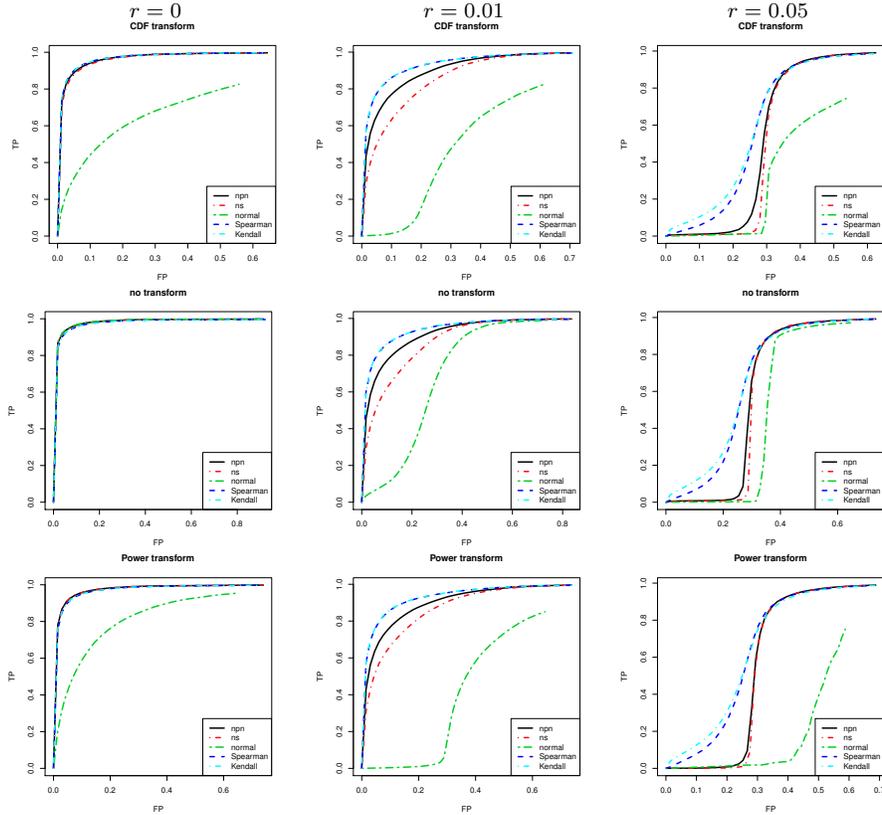

FIG 1. *ROC curves for the cdf, linear and power transformations (top, middle, bottom) using the Meinshausen-Bühlmann graph estimator, with deterministic data contamination at different levels (r=0, 0.01, 0.05). Here $n = 200$ and $d = 100$. Note: "npn" is the original Winsorized normal-score nonparanormal estimator from Liu, Lafferty and Wasserman (2009); "normal" is the naive Gaussian graph estimator; "Spearman" represents the nonparanormal* SKEPTIC *using Spearman's rho; "Kendall" represents the nonparanormal* SKEPTIC *using Kendall's tau; "npn-ns" represents the normal-score based nonparanormal estimator.*

parametric methods (the graphical lasso and Meinshausen-Bühlmann procedure).

From Figures 5 to 8, we see that in terms of oracle FPR and FNR, npn-ns and npn seem slightly better than npn-spearman and npn-tau.

5.2.2. *Non-Gaussian Data with Low Level of Outliers.* When the outlier contamination level is low ($r = 0.01$ for the deterministic contamination and $r = 0.1$ for the random contamination), the performance of the nonparanormal SKEPTIC (npn-spearman and npn-tau) is significantly better than that of npn and npn-ns. Still, all the semiparametric methods significantly outperform the corresponding parametric methods (the graphical lasso and parallel lasso procedure). Similar patterns can also be found based on the quantitative comparisons in Figures 5 to 8.

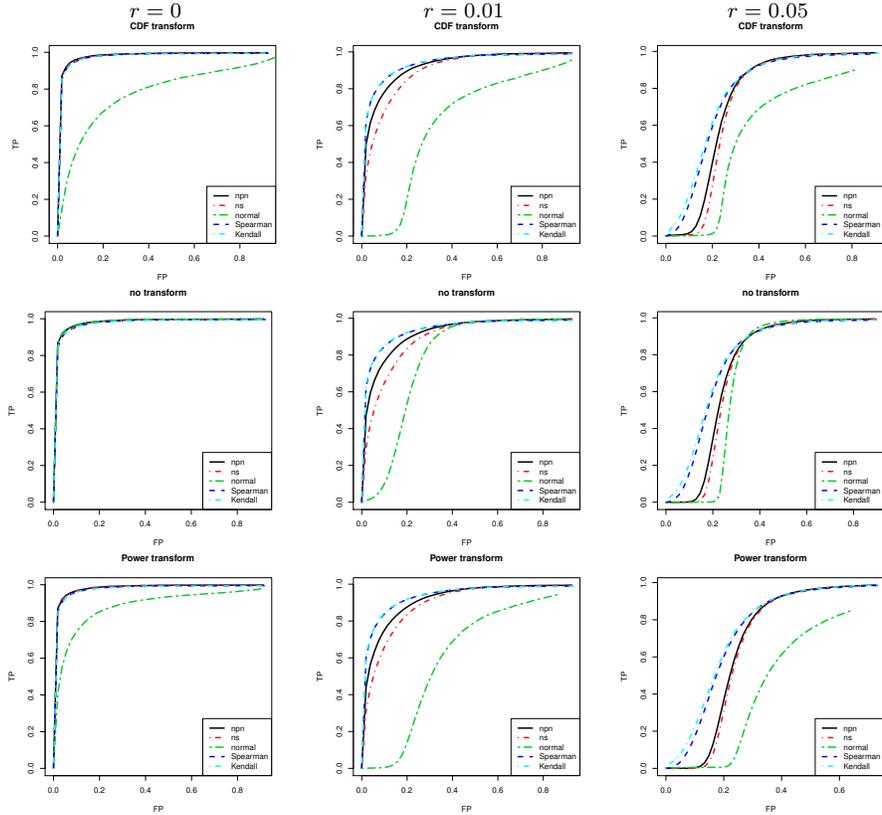

Fig 2. *ROC curves for the cdf, linear and power transformations (top, middle, bottom) using the glasso graph estimator, with deterministic data contamination at different levels (r=0, 0.01, 0.05), with $n = 200$ and $d = 100$.*

5.2.3. *Non-Gaussian Data with Higher Level of Outliers.* From Figures 1 to 4, we see that when the data contamination level is higher ($r = 0.05$ for the deterministic contamination and $r = 0.20$ for the random contamination), the performance of the nonparanormal SKEPTIC (npn-spearman and npn-tau) is significantly better than that of npn and npn-ns. For this high outlier case, npn-tau outperforms npn-spearman, suggesting that Kendall's tau is more robust than Spearman's rho statistic. The parametric methods (the graphical lasso and parallel lasso procedure) perform the worst.

Unlike the previous low outlier case, the quantitative results from Figures 5 to 8 present interesting patterns. For deterministic contamination, we do not see significant improvement of the npn-spearman and npn-tau over npn and npn-ns in terms of the oracle FPR and FNR. At the first sight this seems counter-intuitive since the corresponding ROC curves suggest that npn-spearman and npn-tau are globally better than npn and npn-ns. The main reason for such a result is that the oracle score point happens to coincide with the intersection point of different ROC curves. On the other hand, for the random contamination setting, we see that the performance of npn-spearman and npn-tau uniformly dominate that of the npn and npn-ns.

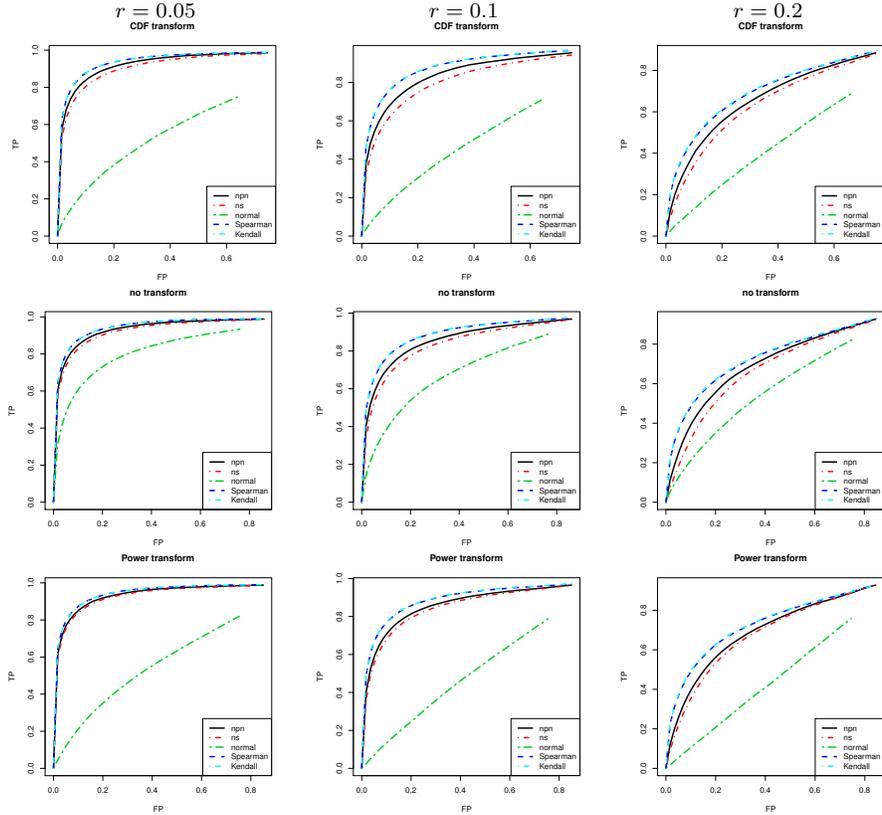

FIG 3. *ROC curves for the cdf, linear and power transformations (top, middle, bottom) using the glasso graph estimator, with random data contamination at different levels (r=0.05, 0.1, 0.2), with $n = 200$ and $d = 100$.*

5.2.4. *Gaussian Data with No Outliers.* From the linear transformation plot in Figures 1 to 4, we see that when the outlier contamination level is $r = 0$ the performance of all these methods are comparable. Based on Figures 5 to 8, we could see that in terms of oracle FPR and FNR, normal, npn-ns and npn are slightly better than npn-spearman and npn-tau. This result suggests that there is only a very small efficiency loss for the nonparanormal SKEPTIC with truly Gaussian data, though this loss seems negligible.

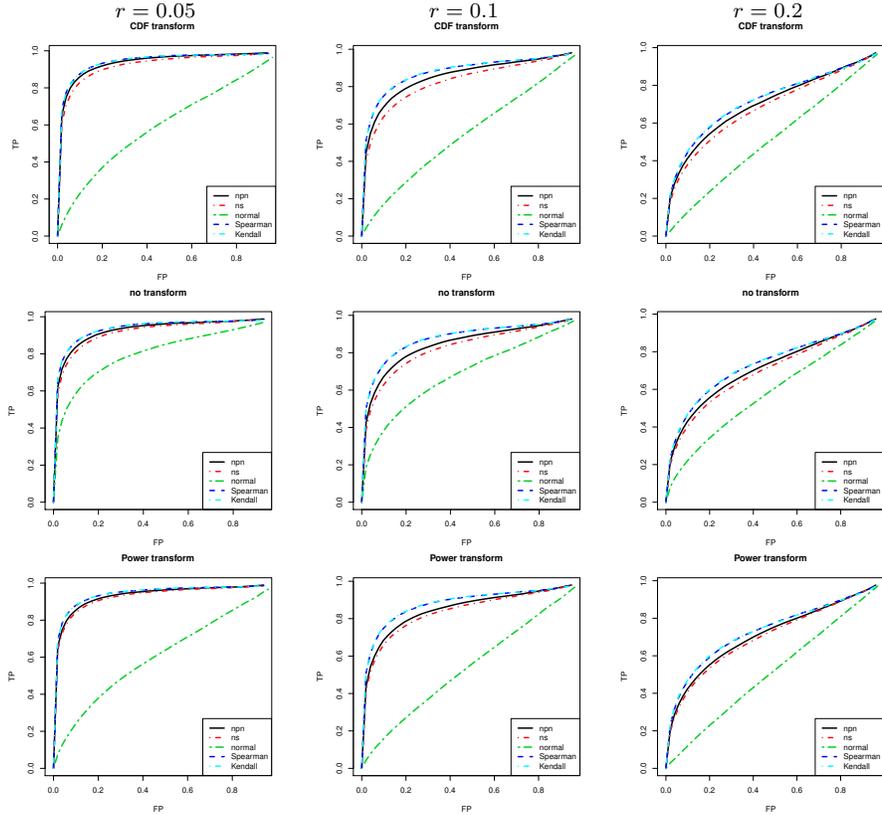

FIG 4. *ROC curves for the cdf, linear and power transformations (top, middle, bottom) using the Meinshausen-Bühlmann graph estimator, with random data contamination at different levels (r=0.05, 0.1, 0.2), with n = 200 and d = 100.*

5.2.5. *Gaussian Data with Low and Higher Levels of Outliers.* From the linear transformation plot in Figures 1 to 4, we see that when the outlier contamination level is $r > 0$, the performance of the parametric methods like the graphical lasso immediately decreases. The main reason is that these methods are based on the Pearson correlation matrix, which is very sensitive to outliers. In contrast, the semiparametric methods (npn-spearman, npn-tau, npn-ns, and npn) are more resistant to outliers. Among them, npn-tau is the most robust, and npn-spearman behaves similarly. Both methods outperform npn, which further outperforms npn-ns.

In summary, the simulation results illustrate an interesting tradeoff between statistical efficiency and estimation robustness. In general, both npn-spearman and npn-tau have very good overall performance. In practice, which method to use should be determined by knowledge about the data. For example, for high-throughput genomics datasets, we believe that using npn-spearman and npn-tau is preferable to using less robust methods like npn-ns.

| tf | r | n | npn | | npn-ns | | normal | | spearman | | kendall | |
|---|---|---|---|---|---|---|---|---|---|---|---|---|
| | | | FPR(%) | FNR | FPR | FNR | FPR | FNR | FPR | FNR | FPR | FNR |
| cdf | 0.00 | 100 | 11(2.9) | 13(3.5) | 11(3.1) | 13(3.6) | 26(6.9) | 38(9.2) | 11(3.4) | 15(3.6) | 11(3.2) | 15(3.6) |
| | | 200 | 6(2) | 5(2.1) | 6(1.9) | 6(2.5) | 18(6.7) | 32(17.2) | 6(2.2) | 6(2.4) | 6(2.1) | 6(2.4) |
| | | 500 | 2(1.6) | 1(1.2) | 3(1.7) | 1(1.1) | 11(4.2) | 19(20.9) | 3(1.6) | 2(1.4) | 3(1.6) | 2(1.4) |
| | 0.01 | 100 | 14(3.8) | 15(3.9) | 16(4.4) | 15(4.5) | 33(8) | 38(11.4) | 13(3.1) | 16(3.8) | 13(3.2) | 16(3.9) |
| | | 200 | 12(3.7) | 16(4.5) | 24(7.8) | 13(6.7) | 40(9.7) | 28(15.8) | 10(2.7) | 12(3.4) | 10(2.8) | 12(3.1) |
| | | 500 | 4(1.6) | 5(2) | 7(2.4) | 8(2.7) | 40(9.3) | 17(14.2) | 3(1.5) | 3(1.5) | 3(1.4) | 3(1.6) |
| | 0.05 | 100 | 27(2.6) | 12(3.5) | 26(2.4) | 12(3.5) | 40(10.4) | 40(13) | 25(2.3) | 14(3.3) | 27(2.9) | 13(3.2) |
| | | 200 | 36(2) | 7(2) | 37(2) | 7(2) | 37(13.8) | 35(24.4) | 36(2.4) | 8(2.5) | 36(2.3) | 8(2.7) |
| | | 500 | 33(1.3) | 1(0.9) | 33(1.2) | 1(1) | 43(10.7) | 21(17.4) | 31(1.4) | 1(1) | 31(1.5) | 1(1.2) |
| linear | 0.00 | 100 | 11(3.2) | 13(3.7) | 11(2.9) | 13(3.1) | 11(2.8) | 12(3.2) | 11(2.6) | 14(3.5) | 11(2.8) | 15(3.5) |
| | | 200 | 6(2.1) | 5(2) | 5(2) | 5(2) | 5(1.5) | 5(4.1) | 6(2) | 6(2.1) | 6(2.1) | 6(2.3) |
| | | 500 | 2(1) | 1(1.1) | 2(1.1) | 1(1) | 2(0.9) | 1(0.7) | 2(0.9) | 1(1.2) | 2(0.9) | 1(1.2) |
| | 0.01 | 100 | 14(3.3) | 16(4.1) | 16(4.3) | 16(4.8) | 25(3.3) | 13(7.6) | 13(3.5) | 16(4) | 13(3.8) | 16(4.5) |
| | | 200 | 13(4.4) | 16(4.6) | 27(5.9) | 11(5.6) | 37(4) | 6(8.2) | 10(2.7) | 12(3.2) | 9(2.9) | 12(3.3) |
| | | 500 | 5(2.1) | 5(2.3) | 7(2.3) | 10(3.4) | 33(2.9) | 2(3.6) | 3(1.2) | 3(1.6) | 3(1.3) | 3(1.6) |
| | 0.05 | 100 | 26(2.4) | 12(3.2) | 27(2.6) | 12(3.3) | 35(4.9) | 17(7.5) | 26(2.4) | 13(3.4) | 27(2.5) | 13(3.1) |
| | | 200 | 37(1.9) | 7(3) | 37(1.9) | 7(2.9) | 37(5.5) | 7(12.1) | 36(2.4) | 8(2.8) | 37(2.6) | 8(2.8) |
| | | 500 | 33(1.4) | 1(1) | 33(1.3) | 1(1.1) | 35(3.3) | 5(5.8) | 31(1.4) | 1(1.1) | 31(1.4) | 1(1.1) |
| power | 0.00 | 100 | 11(2.9) | 13(3.4) | 11(3.2) | 13(3.4) | 25(5) | 32(6.7) | 11(3.3) | 14(3.6) | 12(3.5) | 14(3.7) |
| | | 200 | 6(2.7) | 5(2.4) | 6(2.9) | 5(2.2) | 19(4.2) | 18(6.4) | 6(2.7) | 6(2.7) | 6(2.6) | 6(2.7) |
| | | 500 | 2(1.5) | 1(1.1) | 2(1.4) | 1(1.1) | 9(2.3) | 8(3) | 2(1.3) | 1(1.3) | 2(1.5) | 1(1.3) |
| | 0.01 | 100 | 14(3.5) | 16(4.4) | 16(3.8) | 16(4.4) | 33(5.2) | 32(6.1) | 13(3.6) | 16(4.2) | 13(3.3) | 16(3.9) |
| | | 200 | 12(3.5) | 17(4.3) | 21(7.2) | 15(7.5) | 50(8.5) | 23(13.1) | 10(2.8) | 12(3.3) | 9(2.7) | 12(3.5) |
| | | 500 | 5(1.6) | 5(2) | 5(1.9) | 7(2.3) | 40(4.5) | 13(6.1) | 3(1.4) | 3(1.4) | 3(1.3) | 3(1.5) |
| | 0.05 | 100 | 26(2.3) | 12(3.1) | 26(2.2) | 12(3.2) | 43(6.3) | 41(8.7) | 25(2.5) | 13(3.4) | 26(2.5) | 13(3.3) |
| | | 200 | 37(2.1) | 8(3.1) | 37(2.1) | 8(3.2) | 48(6.8) | 27(11.9) | 36(2.5) | 8(2.8) | 37(2.7) | 8(3.3) |
| | | 500 | 33(1.4) | 1(1.1) | 33(1.2) | 1(1.8) | 47(3.4) | 14(5.3) | 31(1.4) | 1(1.2) | 31(2.8) | 1(3.2) |

FIG 5. *Quantitative comparison of the 5 methods on simulated datasets using different nonparanormal transformations. The graphs are estimated using the glasso algorithm with deterministic data contamination.*

In contrast, if the data are free from outliers, a normal-score based method like npn could be a good choice.

5.3. *Gene Expression Data.* We compare different methods on a large genomics dataset. In this study, we collect 13,182 publicly available microarray samples for Affymetrix's HGU133a platform. These samples are downloaded from GEO and Array Express. Our data contains 2,717 tissue types (e.g., lung cancer, stem cell etc.). For each array sample, there are 22,283 probes, corresponding to 12,719 genes. This is thus far the largest microarray gene expression dataset that has been collected.

The main purpose of this study is to estimate the conditional independence graphs over different genes and different tissue types. To estimate the gene graph, we treat the 13,182 arrays as independent observations and the expression value of each gene as a random variable. To estimate the tissue graph, we average all the arrays belonging to the same tissue type and treat this tissue type expression as a random variable. In this setting, the 12,719 gene expressions are treated as independent observations. While the gene and tissue types are not independent, we adopt this approach as our working procedure, for simplicity.

| tf | $r$ | $n$ | npn | | npn-ns | | normal | | spearman | | kendall | |
|---|---|---|---|---|---|---|---|---|---|---|---|---|
| | | | FPR(%) | FNR | FPR | FNR | FPR | FNR | FPR | FNR | FPR | FNR |
| cdf | 0.00 | 100 | 10(2.8) | 15(4.2) | 10(2.9) | 15(4.4) | 25(5.5) | 44(6.4) | 11(2.6) | 16(4.4) | 11(2.7) | 16(4.4) |
| | | 200 | 4(1.5) | 5(2.5) | 5(1.7) | 6(3) | 20(4.6) | 30(5.4) | 5(1.7) | 5(2.6) | 5(1.9) | 5(2.4) |
| | | 500 | 1(0.7) | 1(0.8) | 1(0.7) | 1(1) | 11(2.9) | 12(3.4) | 1(0.6) | 1(0.9) | 1(0.6) | 1(0.8) |
| | 0.01 | 100 | 12(3.5) | 16(4) | 14(3.3) | 15(3.5) | 33(7.4) | 43(8) | 11(3) | 17(3.9) | 12(3.1) | 16(3.9) |
| | | 200 | 15(3.4) | 12(3.5) | 21(3.4) | 12(3.6) | 38(4.6) | 29(5.1) | 10(3.3) | 13(3.6) | 10(3.1) | 12(3.4) |
| | | 500 | 4(1.7) | 4(2.9) | 6(2.4) | 5(3.3) | 39(3.4) | 14(4.6) | 2(1.4) | 2(2.2) | 2(1.2) | 2(2.2) |
| | 0.05 | 100 | 22(2.5) | 14(3.3) | 23(2.5) | 15(3.5) | 39(7) | 43(7.9) | 21(3.2) | 16(4.1) | 22(3) | 16(4.2) |
| | | 200 | 35(2.8) | 9(3.5) | 35(3) | 9(3.5) | 42(4.3) | 28(5.7) | 32(3.2) | 11(4.1) | 33(3.5) | 11(3.8) |
| | | 500 | 27(2.3) | 3(1.9) | 29(1.9) | 3(1.9) | 46(4.2) | 15(4.6) | 21(2.7) | 4(2.3) | 20(2.6) | 4(2.4) |
| linear | 0.00 | 100 | 10(2.8) | 15(3.5) | 10(2.7) | 14(3.4) | 9(2.5) | 14(3.2) | 11(2.8) | 16(3.6) | 11(2.6) | 16(3.4) |
| | | 200 | 4(1.5) | 5(1.9) | 4(1.5) | 5(1.8) | 4(1.6) | 5(2) | 5(1.5) | 6(2.4) | 5(1.6) | 6(2.3) |
| | | 500 | 1(0.6) | 1(1.1) | 1(0.6) | 1(1.1) | 1(0.6) | 1(1.1) | 1(0.6) | 1(1.3) | 1(0.6) | 1(1.3) |
| | 0.01 | 100 | 12(2.9) | 16(3.9) | 14(3.5) | 16(4.1) | 22(3) | 15(3.7) | 12(3.5) | 17(4) | 11(3.1) | 18(4.2) |
| | | 200 | 16(3.8) | 13(4.3) | 23(3.7) | 11(4.1) | 34(2.3) | 7(2.7) | 10(3.4) | 13(4) | 10(3.1) | 13(3.8) |
| | | 500 | 4(1.5) | 4(1.9) | 7(2.2) | 5(2.2) | 22(2.4) | 4(2.2) | 2(1.1) | 2(1.4) | 2(1) | 2(1.5) |
| | 0.05 | 100 | 23(2.8) | 15(3.3) | 23(2.5) | 15(3.6) | 30(3.9) | 20(4.1) | 22(3.1) | 16(4.1) | 21(3.3) | 17(3.6) |
| | | 200 | 35(2.6) | 9(3.2) | 36(2.6) | 8(3.1) | 37(2.1) | 6(2.2) | 32(2.9) | 10(3.4) | 33(3) | 10(3.3) |
| | | 500 | 27(2.1) | 2(1.5) | 29(1.9) | 2(1.5) | 33(2) | 4(1.8) | 21(2.5) | 4(2.1) | 20(2.7) | 4(2.3) |
| power | 0.00 | 100 | 10(2.9) | 15(3.8) | 10(2.9) | 14(3.9) | 18(4.2) | 33(5.3) | 11(3.1) | 16(4.2) | 10(3.3) | 17(4.2) |
| | | 200 | 4(1.6) | 5(1.9) | 4(1.7) | 5(1.9) | 14(2.9) | 18(4.1) | 5(1.5) | 6(2.2) | 5(1.6) | 6(2.2) |
| | | 500 | 1(0.6) | 1(0.7) | 1(0.5) | 1(0.7) | 7(1.8) | 6(2) | 1(0.5) | 1(0.8) | 1(0.6) | 1(0.7) |
| | 0.01 | 100 | 13(2.9) | 16(3.9) | 14(2.9) | 16(4.4) | 26(5.5) | 37(6.7) | 12(2.8) | 18(3.9) | 12(3) | 17(3.3) |
| | | 200 | 17(4) | 13(4.6) | 21(4) | 12(4.2) | 45(4.6) | 23(5.7) | 11(3.1) | 13(3.8) | 10(3.3) | 13(3.9) |
| | | 500 | 4(1.5) | 4(2.4) | 5(2.1) | 5(2.8) | 36(4.2) | 13(6.4) | 2(1.1) | 2(1.9) | 2(1.4) | 2(2) |
| | 0.05 | 100 | 22(2.8) | 15(3.3) | 23(2.5) | 15(3.3) | 41(9.8) | 42(11) | 20(2.9) | 17(3.6) | 22(2.9) | 17(3.6) |
| | | 200 | 35(2.8) | 9(4.1) | 35(2.6) | 9(3.9) | 50(5.4) | 24(7.5) | 32(2.9) | 10(3.4) | 33(2.9) | 10(3.9) |
| | | 500 | 27(1.9) | 2(1.7) | 28(2.1) | 2(1.7) | 45(3.7) | 14(4.4) | 20(2.4) | 4(2.3) | 20(2.8) | 4(2.5) |

FIG 6. *Quantitative comparison of the 5 methods on simulated datasets using different non-paranormal transformations. The graphs are estimated using the Meinshausen-Bühlmann algorithm with deterministic data contamination.*

Two major challenges for conducting statistical analysis on large-scale integrated datasets are data cleaning and batch/lab effects removal. We conduct surrogate variable analysis (Leek and Storey, 2007) on this data to remove batch effects and normalize the data from different labs. Since the main purpose of this paper is to compare different methods on empirical datasets, we focus on presenting the differential graphs between different methods. The detailed data preprocessing protocols and the scientific implications of the obtained results are not reported.

| | | | npn | | npn-ns | | normal | | spearman | | kendall | |
|---|---|---|---|---|---|---|---|---|---|---|---|---|
| tf | r | n | FPR(%) | FNR | FPR | FNR | FPR | FNR | FPR | FNR | FPR | FNR |
| cdf | 0.05 | 100 | 16(3.6) | 24(4.9) | 17(4.4) | 26(5.7) | 27(12.9) | 57(13.3) | 16(3.9) | 23(4.8) | 16(4.1) | 23(5) |
| | | 200 | 10(2.2) | 12(3) | 11(2.6) | 14(3.6) | 26(10.9) | 51(12.5) | 10(2.8) | 11(3.2) | 9(2.6) | 11(3.3) |
| | | 500 | 4(2.1) | 4(2.5) | 5(2.1) | 6(2.7) | 22(8.3) | 40(13.9) | 4(2.1) | 4(2.2) | 4(2) | 4(2.1) |
| | 0.10 | 100 | 19(5) | 35(6.2) | 20(4.9) | 37(6.3) | 30(17.4) | 59(18) | 17(4.8) | 33(6.1) | 18(4.8) | 33(6.2) |
| | | 200 | 15(3.8) | 21(4.6) | 16(3.9) | 25(5.1) | 29(13.2) | 56(13.3) | 13(3.3) | 18(4.6) | 13(3.5) | 18(4.5) |
| | | 500 | 7(2.3) | 9(2.7) | 9(2.4) | 12(3.1) | 27(11.3) | 50(13) | 6(1.9) | 7(2.2) | 6(2.1) | 6(2.2) |
| | 0.20 | 100 | 28(7.9) | 47(8.2) | 29(7.5) | 48(8.2) | 30(19.2) | 64(20.4) | 24(7.8) | 50(8.2) | 24(7.9) | 49(7.8) |
| | | 200 | 24(6.7) | 39(7.5) | 28(6.7) | 39(6.9) | 31(17.8) | 61(18.6) | 20(5.8) | 37(6.7) | 19(5.7) | 37(6) |
| | | 500 | 17(3.5) | 23(4.6) | 20(4.7) | 28(5) | 34(15.4) | 54(15.6) | 13(3.6) | 19(4.4) | 12(3.3) | 19(4.2) |
| linear | 0.05 | 100 | 15(3.5) | 25(4.6) | 16(4.6) | 26(4.7) | 23(6.3) | 38(6.7) | 15(3.6) | 23(4.6) | 14(3.2) | 24(4.6) |
| | | 500 | 5(2.4) | 4(1.9) | 5(2.4) | 5(2) | 10(2.7) | 12(3.7) | 4(2.2) | 3(1.7) | 4(2.2) | 3(1.6) |
| | | 200 | 10(2.3) | 13(3.4) | 11(2.5) | 14(3.4) | 16(4.3) | 27(8.4) | 9(2.5) | 11(3) | 9(2.2) | 11(3.2) |
| | 0.10 | 100 | 19(4.8) | 35(6) | 20(5.4) | 37(6.3) | 28(10.2) | 48(9.6) | 19(4.6) | 32(5.2) | 18(4.6) | 32(5.3) |
| | | 200 | 14(4) | 22(4.5) | 15(3.8) | 25(4.2) | 24(6.5) | 40(7.1) | 13(3) | 18(4.2) | 12(3.1) | 18(4.3) |
| | | 500 | 8(2.1) | 9(2.7) | 10(2.5) | 11(3.2) | 19(4.6) | 24(4.8) | 6(1.9) | 7(2.4) | 6(2.2) | 6(2.3) |
| | 0.20 | 100 | 28(7.6) | 48(7.8) | 30(9) | 47(8.8) | 35(18) | 53(17.5) | 24(7.6) | 49(7.6) | 24(7) | 49(7.2) |
| | | 200 | 25(5.1) | 37(6.5) | 30(6.5) | 36(7) | 32(11.4) | 50(11.6) | 19(5.3) | 37(6.3) | 18(4.8) | 38(5.7) |
| | | 500 | 18(4) | 23(5.2) | 22(4.8) | 25(5.4) | 27(7.4) | 41(8.2) | 13(3.8) | 19(4.2) | 13(3.5) | 19(4.2) |
| power | 0.05 | 100 | 15(4.5) | 25(5.7) | 16(4.4) | 25(5) | 33(13.2) | 55(13.9) | 15(4.1) | 23(4.8) | 16(4.3) | 22(5.1) |
| | | 200 | 10(3.2) | 13(3.7) | 10(3.1) | 14(3.5) | 30(8.4) | 52(8.9) | 9(2.8) | 12(3.4) | 9(2.7) | 11(3.2) |
| | | 500 | 4(2.2) | 4(1.8) | 5(2) | 5(1.9) | 28(6.9) | 39(8.1) | 4(2) | 3(1.7) | 4(2.1) | 3(1.7) |
| | 0.10 | 100 | 20(4.9) | 35(5.7) | 20(6) | 36(6.4) | 38(22.2) | 56(22.5) | 18(5.2) | 32(5.7) | 18(5.1) | 32(5.8) |
| | | 200 | 14(4.1) | 22(5.2) | 16(3.8) | 23(5.1) | 39(16.4) | 52(17.3) | 13(3.9) | 19(4.5) | 12(3.7) | 18(4.1) |
| | | 500 | 7(2.2) | 9(2.7) | 8(2.2) | 10(2.9) | 37(11.7) | 46(12.1) | 6(1.7) | 6(2.2) | 5(1.7) | 6(2.1) |
| | 0.20 | 100 | 27(7.7) | 48(9.5) | 30(8.4) | 47(9.9) | 42(24.8) | 54(25.6) | 22(7.3) | 50(8.9) | 23(8) | 49(9.2) |
| | | 200 | 24(6) | 38(7.2) | 27(5.9) | 38(7.3) | 41(24.4) | 54(25) | 20(4.7) | 37(5.5) | 19(5.1) | 36(5.8) |
| | | 500 | 18(4) | 23(4.8) | 20(4.2) | 24(5.3) | 41(16.9) | 51(17.7) | 13(3.6) | 19(4.3) | 12(3.1) | 19(4.3) |

Fig 7. *Quantitative comparison of the 5 methods on simulated datasets using different nonparanormal transformations. The graphs are estimated using the glasso algorithm with random data contamination.*

We first screen out all the genes whose marginal standard deviation is below a given threshold. Such a procedure provides us a list of 2,000 genes which vary the most across different array samples. To estimate the gene graph, we first calculate the full regularization path for 100 tuning parameters using npn-spearman and automatically select the tuning parameter using the StARS stability based approach (Liu, Roeder and Wasserman, 2010). The resulting graph contains 1,557 edges. We then examine the full regularization paths of the other graph estimation methods and select the graph with closest sparsity level.

To estimate the tissue network, we first remove all the data for tissue types which have less than 5 replications, leaving 2,714 tissue types. We only use the 2,000 filtered out genes to estimate the tissue network. After averaging the array samples belonging to the same tissue type, we obtain a final data matrix with size 2,000 × 2,714. The remaining procedure of estimating the tissue graph is the same as that of estimating the gene graph. Some summary statistics of the estimated gene and tissue graphs are presented in Figure 9.

| tf | $r$ | $n$ | npn | | npn-ns | | normal | | spearman | | kendall | |
|----|-----|-----|--------|-----|--------|-----|--------|-----|----------|-----|---------|-----|
| | | | FPR(%) | FNR | FPR | FNR | FPR | FNR | FPR | FNR | FPR | FNR |
| cdf | 0.05 | 100 | 15(3.7) | 27(4.3) | 15(3.5) | 30(4.5) | 29(16.1) | 60(15.8) | 13(3.3) | 27(4.4) | 14(3.2) | 26(4.3) |
| | | 200 | 9(2.4) | 13(3.1) | 10(2.7) | 15(4.1) | 27(9.7) | 53(10.5) | 9(2.5) | 11(3.4) | 8(2.7) | 11(3.3) |
| | | 500 | 3(1.5) | 4(1.8) | 4(1.4) | 5(2.2) | 21(5.7) | 42(6.8) | 3(1.3) | 3(1.8) | 3(1.2) | 3(1.8) |
| | 0.10 | 100 | 18(4.7) | 40(5.4) | 18(5.7) | 42(6.6) | 38(21.6) | 55(21.7) | 18(5) | 37(5.8) | 17(5.1) | 36(5.6) |
| | | 200 | 13(3.6) | 25(5.3) | 15(3.9) | 28(5.6) | 32(14.2) | 56(14) | 12(3.2) | 21(5.2) | 12(3.2) | 21(5) |
| | | 500 | 7(2.4) | 10(2.9) | 9(2.9) | 14(3.4) | 24(9.2) | 53(10.6) | 5(1.8) | 6(2.6) | 5(1.5) | 6(2.6) |
| | 0.20 | 100 | 22(8.2) | 55(8.2) | 22(7.8) | 56(8.4) | 50(31.4) | 45(31) | 22(7.7) | 54(9) | 22(7) | 53(8) |
| | | 200 | 16(6.5) | 45(7.5) | 19(7.2) | 48(8) | 36(23.7) | 57(23.5) | 19(6.2) | 40(7.3) | 19(5.5) | 41(7.1) |
| | | 500 | 14(4.1) | 28(5) | 15(3.9) | 35(5.6) | 29(16.3) | 57(15.7) | 12(3) | 21(4.4) | 12(3.4) | 21(4.6) |
| linear | 0.05 | 100 | 14(3.6) | 29(4.9) | 14(3.6) | 30(4.7) | 19(5.8) | 45(6.8) | 14(4) | 26(5.3) | 13(4.3) | 26(5.2) |
| | | 200 | 10(2.9) | 14(3.5) | 10(2.9) | 16(4.2) | 15(4.4) | 31(5) | 9(2.7) | 12(3.1) | 8(2.4) | 12(2.9) |
| | | 500 | 3(1.3) | 3(1.6) | 4(1.5) | 4(1.9) | 8(2.7) | 14(3.3) | 3(1.2) | 3(1.7) | 3(1.1) | 3(1.6) |
| | 0.10 | 100 | 17(5) | 41(6.3) | 17(4.6) | 43(6.2) | 20(6.9) | 59(7.9) | 18(5.2) | 37(6.3) | 18(4.6) | 35(5.8) |
| | | 200 | 14(3.8) | 25(5.2) | 14(4.2) | 29(5.6) | 19(6.6) | 47(6.9) | 12(3.1) | 21(4.4) | 12(3.2) | 21(4.6) |
| | | 500 | 7(2.2) | 10(2.9) | 8(2.6) | 13(3.2) | 14(4.2) | 30(5.8) | 5(1.7) | 7(2.4) | 5(1.7) | 7(2.5) |
| | 0.20 | 100 | 23(9.1) | 54(9.3) | 22(8.8) | 56(9.2) | 28(18) | 61(18.1) | 22(8.4) | 53(8.4) | 23(8.6) | 52(8.8) |
| | | 200 | 19(5.8) | 44(6.7) | 19(5.9) | 47(6.6) | 23(10) | 60(10.2) | 19(5.7) | 40(7) | 19(6) | 39(7.5) |
| | | 500 | 14(3.9) | 29(4.9) | 14(4.2) | 33(6) | 20(7.1) | 48(8.4) | 13(3.7) | 20(4.5) | 12(3.2) | 20(4.2) |
| power | 0.05 | 100 | 15(4.2) | 28(4.9) | 15(3.9) | 29(5) | 30(13.7) | 58(14.4) | 14(4.3) | 26(5.1) | 15(4) | 25(4.8) |
| | | 200 | 9(2.5) | 14(3.9) | 9(2.6) | 15(3.9) | 27(10.4) | 52(10.2) | 8(2.6) | 12(3.2) | 8(2.2) | 12(3.1) |
| | | 500 | 3(1.3) | 3(1.5) | 3(1.3) | 4(1.6) | 20(6.2) | 44(7.2) | 3(1.1) | 2(1.4) | 2(1) | 2(1.3) |
| | 0.10 | 100 | 18(5.2) | 40(5.1) | 18(5.4) | 42(5.6) | 41(25.4) | 52(25) | 17(5) | 37(5.8) | 17(4.8) | 36(5.1) |
| | | 200 | 14(3.9) | 25(5.1) | 14(3.9) | 27(5.6) | 33(20) | 57(19.5) | 12(2.7) | 20(4.4) | 12(3.4) | 20(4.3) |
| | | 500 | 7(1.9) | 10(2.9) | 7(2.3) | 11(3) | 26(11.3) | 55(13) | 5(1.7) | 7(2.2) | 5(1.6) | 6(2.1) |
| | 0.20 | 100 | 22(6.9) | 55(8.4) | 22(7.4) | 56(8.7) | 46(26.9) | 48(26.9) | 21(7.4) | 54(8.3) | 22(7.2) | 52(8.4) |
| | | 200 | 19(5.9) | 44(7.1) | 19(6.4) | 46(7.3) | 43(25.5) | 51(25.5) | 19(6.1) | 40(7.2) | 18(4.9) | 40(6.2) |
| | | 500 | 13(4.1) | 27(5.7) | 14(4.8) | 29(5.7) | 35(18.6) | 56(19.3) | 13(3.4) | 20(4.7) | 12(3.4) | 19(4.5) |

FIG 8. *Quantitative comparison of the 5 methods on simulated datasets using different non-paranormal transformations. The graphs are estimated using the Meinshausen-Bühlmann algorithm with random data contamination.*

From Figure 9, we see that the estimated tissue graph is more dense than the gene graph. Since both graphs contain around 2,000 nodes with more than 1,500 edges, it is not very informative to visualize the whole graphs. Instead we are interested in understanding the differential graphs.

For example, at the gene level, the npn-sp graph contains 1,235 edges that are not in the normal graph. In contrast, the normal graph contains 1,228

| Network | dim | Edge No. | | | Edge diff | | | |
|---|---|---|---|---|---|---|---|---|
| | | spearman | normal | npn-ns | SP > GA | SP < GA | SP > NS | SP < NS |
| Tissue | 2714 | 2639 | 2379 | 2478 | 602 | 342 | 307 | 146 |
| Gene | 2000 | 1557 | 1550 | 1411 | 1235 | 1228 | 691 | 545 |

FIG 9. *Summary statistics of the HGU133a data networks estimated at the gene and tissue levels. Note: GA:= normal; SP:=spearman; NS:= npn-ns. A > B means the number of edges only appear in the estimated graph of A, but not in that of B; A < B is vice versa.*

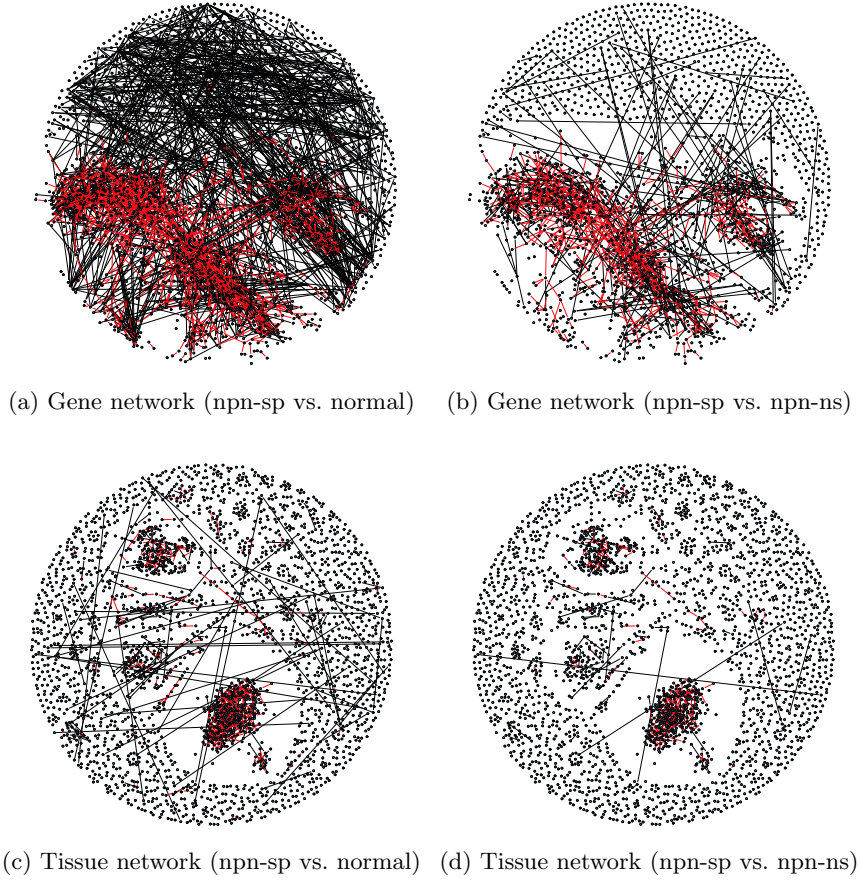

(a) Gene network (npn-sp vs. normal)    (b) Gene network (npn-sp vs. npn-ns)

(c) Tissue network (npn-sp vs. normal)  (d) Tissue network (npn-sp vs. npn-ns)

FIG 10. *Differential gene networks between different methods. For A vs. B, the red color represent the edges that only present in A but not in B, the black color represent the edges that only present in B but not in A. (These graphics are best visualized in color).*

edges that are not in the `npn-sp` graph. Since there are $1{,}235/1{,}557 \approx 80\%$ edges in `npn-sp` that are not present in the `normal` graph, this suggests that the data are highly non-Gaussian. When we further compare the `npn-sp` gene graph with the `npn-ns` graph, we found that there are $691/1{,}557 \approx 45\%$ edges that are not present in the `npn-ns` graph, suggesting that this data may contain high levels of outliers. Since this dataset is integrated from many sources, this is not surprising. Compared with the gene graphs, the tissue graphs present a different pattern. Even though the delivered tissue graphs are much denser than the gene graphs, there are only $602/2{,}714 \approx 22\%$ `npn-sp` edges that are not present in the `normal` graph. Also, there are only $342/2{,}639 \approx 12\%$ edges in the `normal` graph that are not in the `npn-sp` graph. Such a result suggests that the data are still non-Gaussian. However, at the tissue level the data seems to contain a much stronger signal than at the gene level. (This may also be caused by possible uninterpreted lab effects.) A similar conclusion can be drawn when we compare the `npn-spearman` tissue graph with the `npn-ns` tissue graph. For better visualization, we plot the differential graphs in Figure 10. These plots show the difference between the estimated graphs and confirm the above analysis.

**6. Conclusions and Discussion.** Most methods for estimating high dimensional undirected graphs rely on the normality assumption. To weaken this overly restrictive parametric assumption, we propose the *nonparanormal* SKEPTIC. This improved estimator obviates the need to explicitly estimate the marginal transformations, and greatly improves the statistical rate of convergence. Our analysis is non-asymptotic, and the obtained rate is minimax optimal over many model classes. The nonparanormal SKEPTIC can thus be used as a safe replacement for Gaussian based estimators, even when the data are truly Gaussian.

**Acknowledgements.** We are grateful to Professor Peter Bühlmann, the Associate Editor and the three referees for their helpful comments and suggestions.

## APPENDIX A: PROOFS OF MAIN RESULTS

### A.1. Proof of Proposition 3.1.

PROOF. The result on $\tau_{jk}$ directly follows from the definition of $\tau_{jk}$.

Here we prove the result holds for $\rho_{jk}$. Since $F_j(X_j) \sim \text{Uniform}[0, 1]$, we have $\rho_{jk} = 12\mathbb{E}\big[F_j(X_j)F_k(X_k)\big] - 3$. We can also easily show that $\mathbb{E}\big[1 - F_j(X_j)(1 - F_k(X_k))\big] = \mathbb{E}\big[F_j(X_j)F_k(X_k)\big]$. Moreover, we have

$$(A.1) \qquad \mathbb{E}\big[F_j(X_j)F_k(X_k)\big]$$

$$= \mathbb{E}\Big[\mathbb{P}\left(X_j^{(2)} < X_j^{(1)} \,|\, X_j^{(1)}\right)\mathbb{P}\left(X_k^{(3)} < X_k^{(1)} \,|\, X_k^{(1)}\right)\Big]$$

$$= \mathbb{E}\Big[\mathbb{E}\left(I(X_j^{(2)} < X_j^{(1)}, X_k^{(3)} < X_j^{(k)}) \,|\, X_j^{(1)}, X_k^{(1)}\right)\Big].$$

Similarly,

$$(A.2) \qquad \mathbb{E}\big[(1 - F_j(X_j))(1 - F_k(X_k))\big]$$

$$= \mathbb{E}\Big[\mathbb{P}\left(X_j^{(2)} > X_j^{(1)} \,|\, X_j^{(1)}\right)\mathbb{P}\left(X_k^{(3)} > X_k^{(1)} \,|\, X_k^{(1)}\right)\Big]$$

$$= \mathbb{E}\Big[\mathbb{E}\left(I(X_j^{(2)} > X_j^{(1)}, X_k^{(3)} > X_k^{(1)}) \,|\, X_j^{(1)}, X_k^{(1)}\right)\Big].$$

Combining (A.1) and (A.2), we obtain

$$\mathbb{E}\big[F_j(X_j)F_k(X_k)\big] = \frac{1}{2}\mathbb{E}\big[F_j(X_j)F_k(X_k)\big] + \frac{1}{2}\mathbb{E}\big[(1 - F_j(X_j))(1 - F_k(X_k))\big]$$

$$= \frac{1}{2}\mathbb{P}\big((X_j^{(1)} - X_j^{(2)})(X_k^{(1)} - X_k^{(3)}) > 0\big)$$

$$= \frac{1}{2}\mathsf{C}(j, 1, 2; k, 1, 3).$$

Therefore, we have $\rho_{jk} = 12\mathbb{E}\big[F_j(X_j)F_k(X_k)\big] - 3 = 3\,(2\mathsf{C}(j, 1, 2; k, 1, 3) - 1) = 3\mathsf{C}(j, 1, 2; k, 1, 3) - 3\mathsf{D}(j, 1, 2; k, 1, 3)$. The last equality follows from the fact that $\mathsf{C}(j, 1, 2; k, 1, 3) = 1 - \mathsf{D}(j, 1, 2; k, 1, 3)$. $\qquad\square$

### A.2. Proof of Theorem 4.1.

PROOF. The main difficulty of this analysis is that Spearman's rho static is over rank variables which depend on all the samples. To handle this issue, we first rewrite the rho-statistic in a different form (see Page 318, Eq (9.21) of Hoeffding (1948))

$$\widehat{\rho}_{jk} = \frac{3}{n^3 - n}\sum_{i=1}^{n}\sum_{s=1}^{n}\sum_{t=1}^{n}\operatorname{sign}\big(x_j^i - x_j^s\big)\big(x_k^i - x_k^t\big).$$

$$= \frac{n-2}{n+1}U_{jk} + \frac{3}{n+1}\widehat{\tau}_{jk},$$

where $\widehat{\tau}_{jk}$ is Kenadall's tau and $U_{jk} = \frac{3}{n(n-1)(n-2)}\sum_{i\neq s\neq t}\operatorname{sign}\big(x_j^i - x_j^s\big)\big(x_k^i - x_k^t\big)$ is a 3rd-order U-statistic with bounded but asymmetric kernel.

Let $0 < \alpha < 1$. Since $\frac{6}{n+1} > \frac{2}{\pi}(1 - \alpha)c\sqrt{\frac{\log d}{n}}$ whenever $n \geq \frac{9\pi^2}{(1-\alpha)^2 c^2 \log d}$, we have

$$\mathbb{P}\left(\sup_{jk}|\widehat{\rho}_{jk} - \mathbb{E}\widehat{\rho}_{jk}| > \frac{2c}{\pi}\sqrt{\frac{\log d}{n}}\right) \leq \underbrace{d^2\mathbb{P}\left(|U_{jk} - \mathbb{E}U_{jk}| > \frac{2\alpha c}{\pi}\sqrt{\frac{\log d}{n}}\right)}_{T_1(\alpha)}.$$

Without loss of generality, we assume $n$ can be divided by 3. Using Hoeffding's inequality with asymmetric kernels (Hoeffding, 1963),

$$
\begin{aligned}
T_1(\alpha) &= d^2 \mathbb{P}\left(|U_{jk} - \mathbb{E}U_{jk}| > \frac{2\alpha c}{\pi}\sqrt{\frac{\log d}{n}}\right) \\
&\leq 2d^2 \exp\left(-\frac{2}{9\pi^2}\alpha^2 c^2 \left\lfloor\frac{n}{3}\right\rfloor \cdot \frac{\log d}{n}\right) \\
&= 2\exp\left(2\log d - \frac{2}{27\pi^2}\alpha^2 c^2 \log d\right).
\end{aligned}
$$

Let $c = \frac{3\sqrt{6}\pi}{\alpha}$. Therefore, whenever $n \geq \frac{1}{6\log d}\left(\frac{\alpha}{1-\alpha}\right)^2$, with probability at least $1 - 2d^{-2}$, we have $\sup_{jk}|\widehat{\rho}_{jk} - \mathbb{E}\widehat{\rho}_{jk}| \leq \frac{6\sqrt{6}}{\alpha}\sqrt{\frac{\log d}{n}}$.

Unlike $\widehat{\tau}_{jk}$ which is an unbiased estimator of $\tau_{jk}$, $\widehat{\rho}_{jk}$ is a biased estimator. To prove the desired result, we apply the following bias equation from Zimmerman, Zumbo and Williams (2003):

$$
\mathbb{E}\widehat{\rho}_{jk} = \frac{6}{\pi(n+1)}\left[\arcsin\left(\Sigma_{jk}^0\right) + (n-2)\arcsin\left(\frac{\Sigma_{jk}^0}{2}\right)\right].
$$

Equivalently, we can write $\Sigma_{jk}^0 = 2 \cdot \sin\left(\frac{\pi}{6}\mathbb{E}\widehat{\rho}_{jk} + a_{jk}\right)$,

where $a_{jk} = \frac{\pi\mathbb{E}\widehat{\rho}_{jk} - 2 \cdot \arcsin\left(\Sigma_{jk}^0\right)}{2(n-2)}$. It is easy to see that $|a_{jk}| \leq \frac{\pi}{n-2}$.

Therefore, for all $n > \frac{6\pi}{t} + 2$ (which implies that $|a_{jk}| \leq \frac{t}{6}$),

$$
\begin{aligned}
&\mathbb{P}\left(\sup_{jk}\left|\widehat{S}_{jk}^\rho - \Sigma_{jk}^0\right| > t\right) \\
&= d^2 \mathbb{P}\left(\left|2\sin\left(\frac{\pi}{6}\widehat{\rho}_{jk}\right) - 2\sin\left(\frac{\pi}{6}\mathbb{E}\widehat{\rho}_{jk} + a_{jk}\right)\right| > t\right) \\
&\leq d^2 \mathbb{P}\left(\left|\widehat{\rho}_{jk} - \mathbb{E}\widehat{\rho}_{jk} - \frac{6}{\pi}a_{jk}\right| > \frac{3}{\pi}t\right) \\
&\leq d^2 \mathbb{P}\left(\left|\widehat{\rho}_{jk} - \mathbb{E}\widehat{\rho}_{jk}\right| > \frac{3}{\pi}t - \left|\frac{6}{\pi}a_{jk}\right|\right) \\
&\leq d^2 \mathbb{P}\left(\left|\widehat{\rho}_{jk} - \mathbb{E}\widehat{\rho}_{jk}\right| > \frac{3}{\pi}t - \frac{1}{\pi}t\right) \\
&= d^2 \mathbb{P}\left(\left|\widehat{\rho}_{jk} - \mathbb{E}\widehat{\rho}_{jk}\right| > \frac{2}{\pi}t\right).
\end{aligned}
$$

We get the desired result by choosing $\alpha = \frac{3\sqrt{6}}{8}$.                    $\square$

### A.3. Proof of Theorem 4.2.

PROOF. It is easy to see that $\widehat{\tau}_{jk}$ is an unbiased estimator of $\tau_{jk}$: $\mathbb{E}\widehat{\tau}_{jk} = \tau_{jk}$. We have

$$
\begin{aligned}
\mathbb{P}\left(\left|\widehat{S}_{jk}^{\tau} - \Sigma_{jk}^0\right| > t\right) &= \mathbb{P}\left(\left|\sin\left(\frac{\pi}{2}\widehat{\tau}_{jk}\right) - \sin\left(\frac{\pi}{2}\tau_{jk}\right)\right| > t\right) \\
&\leq \mathbb{P}\left(\left|\widehat{\tau}_{jk} - \tau_{jk}\right| > \frac{2}{\pi}t\right).
\end{aligned}
$$

Since $\widehat{\tau}_{jk}$ can be written as a U-statistic: $\widehat{\tau}_{jk} = \frac{2}{n(n-1)}\sum_{1\leq i < i' \leq n} K_{\tau}(x^i, x^{i'})$, where $K_{\tau}(x^i, x^{i'}) = \text{sign}\left(x_j^i - x_j^{i'}\right)\left(x_k^i - x_k^{i'}\right)$ is a kernel bounded between $-1$ and $1$. Using Hoeffding's inequality for U-statistic, we get

$$
\mathbb{P}\left(\sup_{j,k}\left|\widehat{S}_{jk}^{\tau} - \Sigma_{jk}^0\right| > t\right) \leq d^2\exp\left(-\frac{nt^2}{2\pi^2}\right).
$$

We then obtain (4.2). □

### A.4. Proof of Theorem 4.6.

We first present some useful lemmas. Let $\Phi(\cdot)$ and $\phi(\cdot)$ be the cumulative distribution function and density function of standard Gaussian. We start with some preliminary lemmas on the almost sure limit of the Gaussian maxima and the standardized empirical processes. Since $g_j = f_j^{-1}$ and $f_j(t) = \Phi^{-1}(F_j(t))$, we have $g_j(u) = F_j^{-1}(\Phi(u))$.

LEMMA A.1. (Pickands (1969)) Let $z^1, \ldots, z^n \sim N(0,1)$, we then have $\liminf_{n\to\infty}\frac{\sup_{1\leq i\leq n} z^i - \sqrt{2\log n}}{\log\log n/\sqrt{2\log n}} = -\frac{1}{2}$ and $\limsup_{n\to\infty}\frac{\sup_{1\leq i\leq n} z^i - \sqrt{2\log n}}{\log\log n/\sqrt{2\log n}} = \frac{1}{2}$ almost surely.

For any $\gamma > 0$ and $0 < \alpha < 1 < \beta \leq \frac{7}{4}(1-\gamma)$, we define sub-intervals $I_{1n} := \left[g_j(0), g_j\left(\sqrt{\alpha\log n}\right)\right]$ and $I_{2n} := \left[g_j\left(\sqrt{\alpha\log n}\right), g_j\left(\sqrt{\beta\log n}\right)\right]$ and $I_{3n} := \left[g_j\left(\sqrt{\beta\log n}\right), g_j\left(\sqrt{\frac{7}{4}(1-\gamma)\log n}\right)\right]$. We also define

$$(A.3) \quad u_n^* := \sqrt{2\log n} - \frac{\log\log n}{\sqrt{2\log n}} \quad \text{and} \quad t_n^* := \sqrt{2\log n} + \frac{\log\log n}{\sqrt{2\log n}}.$$

LEMMA A.2. For all $t \in I_{1n} \cup I_{2n} \cup I_{3n}$, we have, for large enough $n$, $\frac{1}{n} \leq \widetilde{F}_j(t) \leq 1 - \frac{1}{n}$ almost surely.

PROOF. By Lemma A.1, for any $c > 0$ and large enough $n$, we have the standard Gaussian random variables $z^1, \ldots, z^n$ satisfy $\sup_{1\leq i\leq n} z^i \in$ $\left[\sqrt{2\log n} - \left(\frac{1}{2} + c\right)\frac{\log\log n}{\sqrt{2\log n}}, \sqrt{2\log n} + \left(\frac{1}{2} + c\right)\frac{\log\log n}{2\sqrt{\log n}}\right]$ almost surely. Let $c = \frac{1}{2}$, we have, for large enough $n$,

$$\mathbb{P}\left(\sup_{1 \le i \le n} z_i \in \left[\sqrt{2\log n} - \frac{\log\log n}{\sqrt{2\log n}}, \sqrt{2\log n} + \frac{\log\log n}{\sqrt{2\log n}}\right]\right) = 1.$$

Using the definitions in (A.3), we have, for large enough $n$, $\sup_{1 \le i \le n} x_j^i \in [g_j(u_n^*), g_j(t_n^*)]$ almost surely. Therefore, $\sup_{1 \le i \le n} x_j^i \notin I_{1n} \cup I_{2n} \cup I_{3n}$ almost surely. From the definition of $\widetilde{F}_j$, only the values greater or equal to the $\sup_{1 \le i \le n} x_j^i$ are truncated. The result then follows.   □

The next lemma is from Chapter 16 of Shorack and Wellner (1986). It characterizes the almost sure limit of the standardized empirical process.

Lemma A.3.   *(Almost Sure Limit of the Standardized Empirical Process) Consider a sequence of sub-intervals $[L_n^{(j)}, U_n^{(j)}]$ with both $L_n^{(j)} = g_j(\sqrt{\alpha\log n}) \uparrow \infty$ and $U_n^{(j)} = g_j(\sqrt{\beta\log n}) \uparrow \infty$, then for $0 < \alpha < \beta \le \frac{7}{4}(1-\gamma)$*

$$\limsup_{n \to \infty} \sqrt{\frac{n}{2\log\log n}} \sup_{L_n^{(j)} < t < U_n^{(j)}} \left| \frac{\widetilde{F}_j(t) - F_j(t)}{\sqrt{F_j(t)(1 - F_j(t))}} \right| = C \quad a.s.$$

*where $0 < C \le 2\sqrt{2}$ is a constant.*

Proof.  This result follows from a combination of Theorem 1 and Theorem 2 (Chapter 16) of Shorack and Wellner (1986).   □

The following lemma characterizes the behavior of a random sequence using a deterministic sequence.

Lemma A.4.   *For any $0 < \alpha < 2$, there exists a constant $C$, such that*

$$\limsup_{n \to \infty} \frac{(\Phi^{-1})'(\max\{\widetilde{F}_j(g_j(\sqrt{\alpha\log n})), F_j(g_j(\sqrt{\alpha\log n}))\})}{(\Phi^{-1})'(F_j(g_j(\sqrt{\alpha\log n})))} \le C \text{ almost surely.}$$

Proof.  It suffices to consider the case $\widetilde{F}_j > F_j$. First, for large enough $n$

$$(A.4) \qquad \sqrt{\frac{\phi(\sqrt{\alpha\log n})}{\sqrt{\alpha\log n}}} \le \phi\left(\sqrt{\alpha\log n} + 4\sqrt{\frac{\log\log n}{n^{1-\alpha/2}}}\right) \cdot n^{\alpha/4}.$$

This is true since $\phi\left(\sqrt{\alpha\log n} + 4\sqrt{\frac{\log\log n}{n^{1-\alpha/2}}}\right) = \phi(\sqrt{\alpha\log n}) \cdot (1 - o(1))$.

Therefore, $\phi\left(\sqrt{\alpha \log n} + 4\sqrt{\frac{\log \log n}{n^{1-\alpha/2}}}\right) \cdot n^{\alpha/4} \geq \frac{n^{-\alpha/4}}{2\sqrt{\pi}}$ and $\sqrt{\frac{\phi(\sqrt{\alpha \log n})}{\sqrt{\alpha \log n}}} = \frac{n^{-\alpha/4}}{(2\pi\alpha \log n)^{1/4}}$. Equation (A.4) follows from a combination of these results.

Further, using the fact that $1 - \Phi(t) \leq \frac{\phi(t)}{t}$ for $t \geq 1$, we have

$$4\sqrt{\frac{\log \log n}{n}}\sqrt{1 - \Phi(\sqrt{\alpha \log n})} \leq 4\sqrt{\frac{\log \log n}{n}}\sqrt{\frac{\phi(\sqrt{\alpha \log n})}{\sqrt{\alpha \log n}}}$$

$$\leq 4 \cdot \phi\left(\sqrt{\alpha \log n} + 4\sqrt{\frac{\log \log n}{n^{1-\alpha/2}}}\right)\sqrt{\frac{\log \log n}{n^{1-\alpha/2}}}$$

$$\leq \Phi\left(\sqrt{\alpha \log n} + 4\sqrt{\frac{\log \log n}{n^{1-\alpha/2}}}\right) - \Phi(\sqrt{\alpha \log n}),$$

where the last step follows from the mean value theorem.

Thus $\Phi(\sqrt{\alpha \log n}) + 4\sqrt{\frac{\log \log n}{n}}\sqrt{1 - \Phi(\sqrt{\alpha \log n})} \leq \Phi\left(\sqrt{\alpha \log n} + 4\sqrt{\frac{\log \log n}{n^{1-\alpha/2}}}\right)$. By applying $\Phi^{-1}(\cdot)$ on both sides and the fact that $F_j(g_j(t)) = \Phi(t)$, we have

$$\Phi^{-1}\left(F_j\left(g_j(\sqrt{\alpha \log n})\right) + 4\sqrt{\frac{\log \log n}{n}}\sqrt{1 - F_j\left(g_j(\sqrt{\alpha \log n})\right)}\right)$$

$$\leq \sqrt{\alpha \log n} + 4\sqrt{\frac{\log \log n}{n^{1-\alpha/2}}}.$$

From Lemma A.3, for large enough $n$, $\widetilde{F}_j(t) \leq F_j(t) + 4\sqrt{\frac{\log \log n}{n}} \cdot \sqrt{1 - F_j(t)}$. Therefore $\Phi^{-1}\left(\widetilde{F}_j\left(g_j(\sqrt{\alpha \log n})\right)\right) \leq \sqrt{\alpha \log n} + 4\sqrt{\frac{\log \log n}{n^{1-\alpha/2}}}$. Finally, we have

$$(\Phi^{-1})'\left(\widetilde{F}_j\left(g_j(\sqrt{\alpha \log n})\right)\right) = \frac{1}{\phi\left(\Phi^{-1}\left(\widetilde{F}_j\left(g_j(\sqrt{\alpha \log n})\right)\right)\right)}$$

$$\leq \sqrt{2\pi}\exp\left(\frac{\left(\sqrt{\alpha \log n} + 4\sqrt{\frac{\log \log n}{n^{1-\alpha/2}}}\right)^2}{2}\right)$$

$$\asymp (\Phi^{-1})'\left(F_j\left(g_j(\sqrt{\alpha \log n})\right)\right).$$

This finishes the proof. □

PROOF. of Theorem 4.6. Due to symmetricity, we only need to conduct analysis on a sub-interval of $I_n^s \subset I_n$: $I_n^s := \left[g_j\left(0\right), g_j\left(\sqrt{\frac{7}{4}(1-\gamma)\log n}\right)\right]$.

Recall that for any $0 < \gamma < 1$ and $0 < \alpha < 1 < \beta \leq \frac{7}{4}(1-\gamma)$, we define $I_{1n} := \left[ g_j(0), g_j\left(\sqrt{\alpha \log n}\right) \right]$ and $I_{2n} := \left[ g_j\left(\sqrt{\alpha \log n}\right), g_j\left(\sqrt{\beta \log n}\right) \right]$ and $I_{3n} := \left[ g_j\left(\sqrt{\beta \log n}\right), g_j\left(\sqrt{\frac{7}{4}(1-\gamma)\log n}\right) \right]$. By Lemma A.2, we know that on $I_{1n} \cup I_{2n} \cup I_{3n}$, $\frac{1}{n} \leq \widetilde{F}_j(t) \leq 1 - \frac{1}{n}$ for large enough $n$ almost surely. Therefore, we only need to analyze the term

$$\sup_{t \in I_{1n} \cup I_{2n} \cup I_{3n}} \left| \Phi^{-1}\left( \widetilde{F}_j(t) \right) - \Phi^{-1}\left( F_j(t) \right) \right|.$$

We first consider the term $\sup_{t \in I_{1n}} \left| \Phi^{-1}\left( \widetilde{F}_j(t) \right) - \Phi^{-1}\left( F_j(t) \right) \right|$. Since $\Phi^{-1}$ is a continuous function on the interval between $\min\{\widetilde{F}_j(g_j(0)), F_j(g_j(0))\}$ and $\max\{\widetilde{F}_j(g_j(\sqrt{\alpha \log n})), F_j(g_j(\sqrt{\alpha \log n}))\}$ and is differentiable on the corresponding open set, by the mean-value theorem, for some $\xi_{n,t}$, such that $\xi_{n,t} \in \left[ \min\{\widetilde{F}_j(g_j(0)), F_j(g_j(0))\}, \max\{\widetilde{F}_j(g_j(\sqrt{\alpha \log n})), F_j(g_j(\sqrt{\alpha \log n}))\} \right]$. Thus, $\left| \Phi^{-1}\left( \widetilde{F}_j(t) \right) - \Phi^{-1}\left( F_j(t) \right) \right| = \left| (\Phi^{-1})'(\xi_{n,t}) \left( \widetilde{F}_j(t) - F_j(t) \right) \right|$ for $t \in I_{1n}$.

By Lemma A.4, the following inequality holds almost surely:

$$
\begin{aligned}
(\Phi^{-1})'(\xi_{n,t}) &\leq (\Phi^{-1})'\left( \max\{ F_j\left( g_j\left( \sqrt{\alpha \log n} \right) \right), \widetilde{F}_j\left( g_j\left( \sqrt{\alpha \log n} \right) \right) \} \right) \\
&\leq C(\Phi^{-1})'\left( F_j\left( g_j\left( \sqrt{\alpha \log n} \right) \right) \right) = \frac{C}{\phi\left( \sqrt{\alpha \log n} \right)} \leq c_1 n^{\alpha/2},
\end{aligned}
$$

where $C$ and $c_1$ are some generic constants and $\phi(\cdot)$ is the standard Gaussian density function.

Using $|(\Phi^{-1})'(\xi_{n,t})| \leq c_1 n^{\alpha/2}$ and the Dvoretzky-Kiefer-Wolfowitz inequality, we have $\sup_{t \in I_{1n}} \left| \Phi^{-1}\left( \widetilde{F}_j(t) \right) - \Phi^{-1}\left( F_j(t) \right) \right| = O_P\left( \sqrt{\frac{\log \log n}{n^{1-\alpha}}} \right)$. Next, we consider the term $\sup_{t \in I_{2n}} \left| \Phi^{-1}\left( \widetilde{F}_j(t) \right) - \Phi^{-1}\left( F_j(t) \right) \right|$. By Lemma A.3, for large enough $n$,

$$
\begin{aligned}
\sup_{t \in I_{2n}} \left| \widetilde{F}_j(t) - F_j(t) \right| &= O_P\left( \sqrt{\frac{\log \log n}{n}} \cdot \sqrt{1 - F_j\left( g_j(\sqrt{\alpha \log n}) \right)} \right) \\
&= O_P\left( \sqrt{\frac{\log \log n}{n}} \cdot \sqrt{\frac{n^{-\alpha/2}}{\sqrt{\alpha \log n}}} \right) = O_P\left( \sqrt{\frac{\log \log n}{n^{\alpha/2+1}}} \right).
\end{aligned}
$$

Similarly, we have $\sup_{t \in I_{2n}} |\Phi^{-1}\left( \widetilde{F}_j(t) \right) - \Phi^{-1}\left( F_j(t) \right)| = O_P\left( \sqrt{\frac{\log \log n}{n^{1+\alpha/2-\beta}}} \right)$ and $\sup_{t \in I_{3n}} |\Phi^{-1}\left( \widetilde{F}_j(t) \right) - \Phi^{-1}\left( F_j(t) \right)| = O_P\left( \sqrt{\frac{\log \log n}{n^{\beta/2-3/4+7\gamma/4}}} \right)$. By choosing $\beta = \frac{3}{2}(1-\gamma)$ and $\alpha = 1 - \gamma$, all terms vanish. $\qquad \square$

HAN LIU
DEPARTMENT OF OPERATIONS RESEARCH
AND FINANCIAL ENGINEERING
PRINCETON UNIVERSITY
PRINCETON, NJ, 08544
E-MAIL: hanliu@princeton.edu

FANG HAN
DEPARTMENT OF BIOSTATISTICS
BLOOMBERG SCHOOL OF PUBLIC HEALTH
JOHNS HOPKINS UNIVERSITY
BALTIMORE, MD, 21205
E-MAIL: fhan@jhsph.edu

MING YUAN
SCHOOL OF INDUSTRIAL AND SYSTEMS ENGINEERING
GEORGIA INSTITUTE OF TECHNOLOGY
NW ATLANTA, GA 30332
E-MAIL: myuan@isye.gatech.edu

JOHN LAFFERTY
DEPARTMENT OF STATISTICS
UNIVERSITY OF CHICAGO
CHICAGO, IL 60637
E-MAIL: lafferty@galton.uchicago.edu

LARRY WASSERMAN
DEPARTMENT OF STATISTICS
CARNEGIE MELLON UNIVERSITY
PITTSBURGH, PA 15213
E-MAIL: larry@stat.cmu.edu